\documentclass{article}
\usepackage[a4paper,margin=1in]{geometry}
\usepackage{graphicx} 
\graphicspath{{Figures/}}
\usepackage{tikz}
\usepackage{amsmath}
\usepackage{amssymb}
\usepackage{mathabx}
\usepackage{comment}
\usepackage{appendix}
\usepackage[authoryear]{natbib}
\bibpunct{(}{)}{,}{a}{}{,}
\usepackage{hyperref}
\hypersetup{colorlinks=true, linkcolor=blue, citecolor=blue, urlcolor=blue}
\usepackage{import}
\usepackage{authblk}
\usepackage{array}

\newcommand{\vecx}{\pmb{x}}
\newcommand{\vecy}{\pmb{y}}
\newcommand{\datax}{\pmb{X}}
\newcommand{\datay}{\pmb{Y}}
\newcommand{\meas}{\pmb{p}_{stacked}}
\newcommand{\lat}{\pmb{\xi}}
\newcommand{\f}{\pmb{f}}
\newcommand{\bo}[1]{\pmb{#1}}

\title{Low-Order Flow Reconstruction and Uncertainty Quantification in Disturbed Aerodynamics Using Sparse Pressure Measurements}
\author[1,*]{Hanieh Mousavi}
\author[2]{Jeff Eldredge}
\affil[1,2]{University of California, Los Angeles, Department of Mechanical and Aerospace Engineering}
\affil[*]{Corresponding author, email: hnmousavi@ucla.edu}
\date{}

\begin{document}

\maketitle

\begin{abstract}
    This paper presents a novel machine-learning framework for reconstructing low-order gust-encounter flow field and lift coefficients from sparse, noisy surface pressure measurements. Our study thoroughly investigates the time-varying response of sensors to gust-airfoil interactions, uncovering valuable insights into optimal sensor placement. To address uncertainties in deep learning predictions, we implement probabilistic regression strategies to model both epistemic and aleatoric uncertainties. Epistemic uncertainty, reflecting the model's confidence in its predictions, is modeled using Monte Carlo dropout—as an approximation to the variational inference in the Bayesian framework, treating the neural network as a stochastic entity. On the other hand, aleatoric uncertainty, arising from noisy input measurements, is captured via learned statistical parameters, which propagates measurement noise through the network into the final predictions. Our results showcase the efficacy of this dual uncertainty quantification strategy in accurately predicting aerodynamic behavior under extreme conditions while maintaining computational efficiency, underscoring its potential to improve online sensor-based flow estimation in real-world applications.\\
\end{abstract}

\section{Introduction}\label{introduction}
Many air vehicles operate in highly unsteady aerodynamic environments, such as gust encounters \citep{jones2022physics}. Estimating transient flow fields and aerodynamic loads from sparse measurements in such scenarios is a complex inverse problem due to disturbed flow fields. Accurate flow and aerodynamic load prediction is critical for aerodynamic control, as it enables the design of robust control systems and adaptive mechanisms for dynamic flow conditions. By accurately reconstructing flow fields and quantifying uncertainties, these estimations enhance sensor-based predictions in gust-encounter scenarios, improving the overall reliability of aerodynamic performance. Traditional Bayesian approaches, such as the Ensemble Kalman Filter and its variants, have been widely used to incorporate uncertainty into predictions \citep{le2021ensemble, le2022low}, but they struggle with high-dimensional state spaces, leading to inefficiency and reduced accuracy. This highlights the need for more robust, data-driven techniques for modeling input-output relationships that can be utilized offline for efficient predictions. Deep learning (DL), known for its ability to learn complex and nonlinear mappings, offers a promising alternative. For instance, \citet{dubois2022machine} utilized both linear and nonlinear neural networks (NNs) to reconstruct velocity fields, while \citet{zhong2023sparse} developed a model using Long-Short Term Memory and transfer learning for aerodynamic force and wake reconstruction. \citet{chen2024sparse} applied a multi-layer perceptron (MLP) to estimate aerodynamic loads from surface pressure measurements. Despite these advances, challenges remain in managing numerous parameters and mitigating computational costs for high-dimensional data.

Modern DL constitutes an incredibly powerful tool for regression and classification tasks, as well as for reinforcement learning, where an agent interacts with the environment and learns to take actions that maximize rewards. Deep learning has garnered tremendous attention from researchers across various fields, including physics, biology, medicine, and engineering \citep{tanaka2021deep, ching2018opportunities, akay2019deep, che2023deep}. Despite their broad applicability, DL models are prone to overfitting \citep{brunton2022data}. Moreover, they tend to be overconfident in their predictions, which is particularly problematic in decision-making applications such as safety-critical systems \citep{le2018uncertainty}, medical diagnosis \citep{laves2019quantifying}, and autonomous driving \citep{shafaei2018uncertainty}. Overconfident predictions can lead to poor decision-making and potentially catastrophic consequences if the model's predictions are trusted without question. Therefore, it is crucial to train uncertainty-aware NNs to mitigate these risks and ensure reliable predictions.

There are generally two main sources of uncertainty in DL, i.e. aleatoric and epistemic uncertainties \citep{hullermeier2021aleatoric}. Aleatoric uncertainty ---also known as data uncertainty--- refers to the irreducible uncertainty in data that gives rise to uncertainty in predictions. This type of uncertainty is due to the randomness and noise inherent in the measurements or observations. Aleatoric uncertainty is intrinsic to the process being studied and cannot be eliminated. In contrast, epistemic uncertainty ---also known as model uncertainty--- arises from the lack of knowledge about the best model to describe the underlying data-generating process. Unlike aleatoric uncertainty, this type of uncertainty can be reduced by gathering more data or improving the model. Various approaches exist to propagate aleatoric uncertainty through artificial neural networks (ANNs). One prevalent method is moment matching \citep{frey1999variational,petersen2024uncertainty} which involves propagating the first two moments of a distribution through the network. However, this method increases the number of learned parameters in the network and adds computational cost, especially for large networks. Researchers have also utilized Variational Autoencoders (VAEs) to extract a stochastic latent space from noisy data \citep{gundersen2021semi,liu2022uncertainty}.

Bayesian probability theory offers a robust framework for addressing model uncertainty. In particular, Bayesian neural networks (BNNs), thoroughly reviewed in \citet{jospin2022hands}, are stochastic neural networks trained using Bayesian inference. BNNs can model both aleatoric and epistemic uncertainties. Aleatoric uncertainty is addressed by learning the parameters of a probability distribution at the last layer that approximates the true distribution \citep{jospin2022hands}. Epistemic uncertainty, on the other hand, is modeled by introducing stochastic weights or activations in the deep learning models. By specifying a prior distribution over these stochastic parameters and defining a likelihood function, the exact posterior distribution can be learned through Bayes' rule using Markov Chain Monte Carlo (MCMC) \citep{salakhutdinov2008bayesian} or approximated with a family of distributions using Variational Inference (VI) \citep{swiatkowski2020k}.

In spite of their clear advantages for modeling uncertainty, BNNs often come with prohibitive computational costs and are challenging to converge for large models. However, we can draw upon key aspects of BNN structure, e.g., learning the parameters of a model distribution, to capture aleatoric uncertainty in an efficient manner. Moreover, \citet{gal2016dropoutbayesianapproximationappendix} and \citet{gal2016dropout} have proved that we can interpret dropout in NN---which is traditionally used to prevent overfitting \citep{srivastava2014dropout}---as a Bayesian approximation of a Gaussian process \citep{williams2006gaussian}, without modifying the models themselves. Monte Carlo dropout, known as \emph{MC dropout}, can be used to estimate the uncertainty of the model \citep{gal2016dropout}. In another study, \citet{kendall2017uncertainties} successfully integrated both aleatoric and epistemic uncertainties into a single computer vision model.

This paper aims to estimate aerodynamic flow fields from sensor measurements while incorporating uncertainty quantification within deep learning models. In particular, we use machine learning tools to reconstruct the flow field and the lift coefficient under extreme aerodynamic conditions from sparse surface pressure measurements. Our approach leverages a nonlinear lift-augmented autoencoder, as proposed by \citet{fukami2023grasping}, which captures low-dimensional representations of complex flow dynamics, for improved sensor-based estimation. In this framework, we rigorously analyze the sensor response to gust-airfoil interactions, providing insight into optimal sensor placement. To further enhance prediction robustness, we introduce novel approaches for modeling uncertainties in deep learning predictions, distinguishing between epistemic (model) and aleatoric (data) uncertainties. Following the methodology of \citet{gal2016dropoutbayesianapproximationappendix} and \citet{gal2016dropout}, we apply MC dropout to treat the network stochastically and capture model uncertainty. To capture data uncertainty, our network is trained to estimate the statistical parameters (moments) of a model distribution in a reduced-order latent space, accounting for the inherent noise in the surface pressure data. Our results demonstrate the efficacy of these methods in quantifying two types of uncertainty in a challenging aerodynamic environment.

The paper is structured as follows: Section \ref{methodology} outlines the problem and details the mathematical approach employed for data compression and uncertainty quantification. Section \ref{results} presents the findings of the study. Finally, Section \ref{conclusion} summarizes the key outcomes and implications of the research.

\section{Problem statement and methodology}\label{methodology}
The present study proposes a framework designed to model the intricate and uncertain relationship between input surface pressure measurements and the resulting aerodynamic forces and vortical structure. Our approach leverages advanced data compression techniques and uncertainty quantification to enhance prediction accuracy and reliability. Specifically, we employ DL models to map input surface measurements to a low-dimensional latent space, facilitating efficient reconstruction of flow fields. The framework leverages MC dropout to model epistemic uncertainty in the NN, while incorporating learned loss attenuation to address measurement noise. This combined approach enables robust quantification of confidence intervals, providing a comprehensive assessment of uncertainty in the predictions.

This section outlines the mathematical framework of our approach, covering the problem formulation, NN architecture, and model training and validation using high-fidelity simulation data. Additionally, we discuss the construction of the latent space and the integration of uncertainty quantification techniques into the predictive model, ensuring accurate and reliable performance.

\subsection{Problem statement}\label{probstatement}
\begin{figure}[ht]
\centering
\includegraphics[width=0.7\textwidth, trim=0 50 0 50, clip]{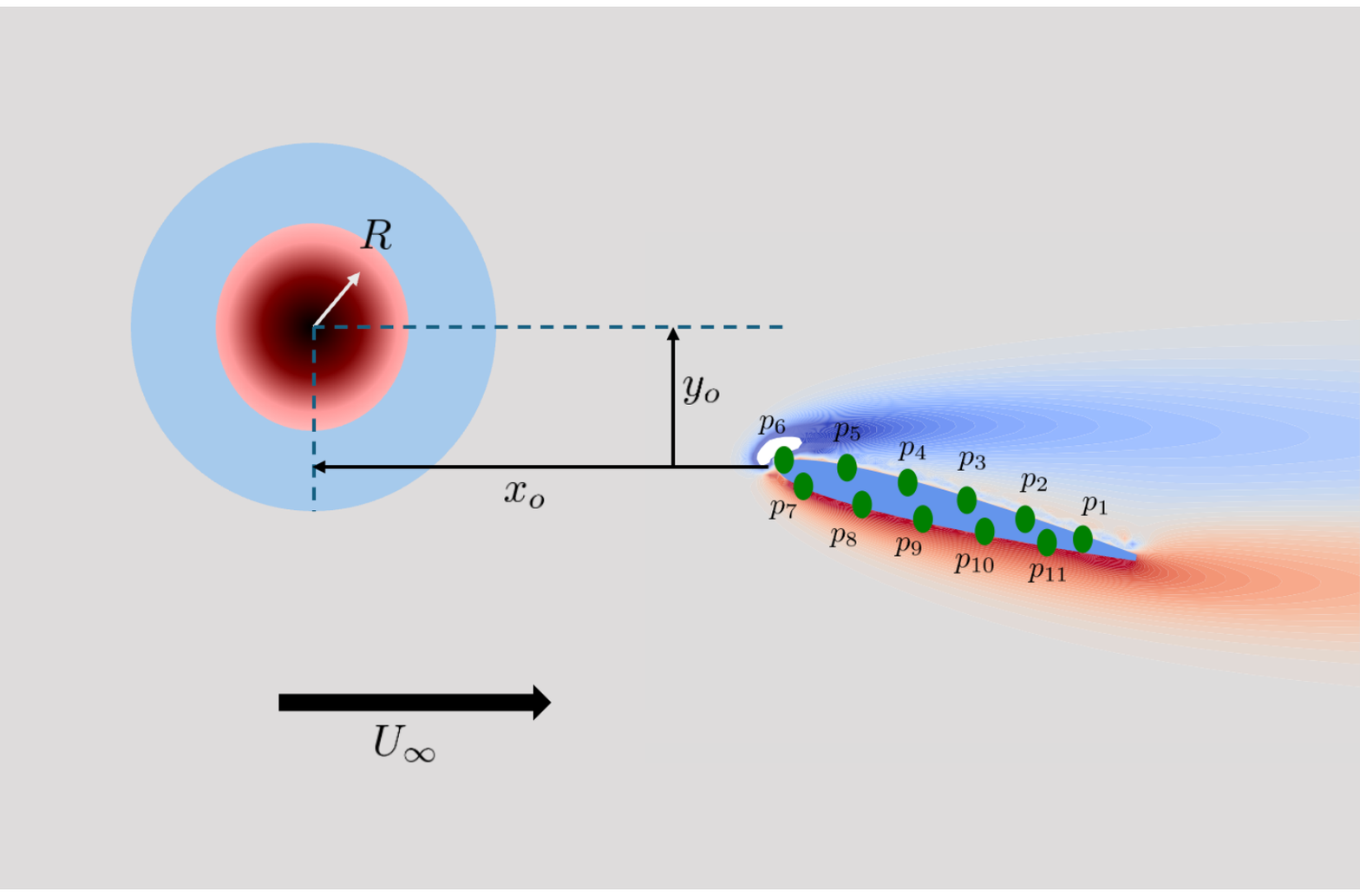}
\caption{\label{fig:gust_approach} Configuration of the problem, illustrating the relative position of the gust center with respect to the airfoil tip, the size of the disturbance, and the indices of sensors mounted on the airfoil.}
\end{figure}

Given sparse pressure measurements from surface sensors, the goal of this work is to estimate the vorticity field and aerodynamic loads from a probabilistic perspective. For data generation in this study, we consider unsteady two-dimensional flow over a NACA 0012 airfoil positioned at a range of angles of attack $\alpha \in \{20^\circ, 30^\circ, $ $40^\circ, 50^\circ, 60^\circ\}$. The freestream velocity is denoted by $U_{\infty}$ with the chord-based Reynolds number $Re=U_\infty c/\nu =100$, where $c$ is the chord length. The case at $\alpha = 20^\circ$ corresponds to a nearly steady flow, while vortex shedding is observed at higher angles of attack. For gust-encounter aerodynamics, the disturbance vortex is modeled as a Taylor vortex \citep{taylor1918dissipation}, 
\begin{equation}
    u_\theta = u_{\theta,max} \frac{r}{R} \exp \left( \frac{1}{2} - \frac{r^2}{2R^2} \right),
\end{equation}
where $R$ is the radius and $u_{\theta,max}$ is the maximum rotational velocity of the vortex. The problem configuration, including the position of sensors and their indices, is illustrated in Fig.~\ref{fig:gust_approach}. The vortex is initially placed upstream of the airfoil at $(x_o,y_o)$ with $x_o/c=-2$, with the origin $(0,0)$ set at the tip of the airfoil. The vortex strength is characterized by $G \equiv u_{\theta,max}/U_{\infty}$. We consider cases with randomly sampled parameters: $G \in [-1,1]$, $y_o/c \in [-0.5,0.5]$, and $2R/c \in [0.5,1]$. The direct numerical simulation of the Navier-Stokes equations in vorticity-streamfunction form is carried out by using the lattice Green's function/immersed layers method proposed by \citet{eldredge2022method} over a domain size of $(-4,4) \times (-2,2)$ on a Cartesian grid with uniform spacing $\Delta x = 0.02$. (The use of the lattice Green's function enables a much tighter domain than other conventional flow techniques.) We numerically calculate the surface pressure $p_s$ relative to ambient pressure $p_\infty$, and define the pressure coefficient for discussion purposes as follows:
\begin{equation}
    C_p = \frac{2(p_s - p_\infty)}{\rho U_\infty^2},
\end{equation}
where $\rho$ represents the fluid density.

Data for the regression task is collected from a portion of the computational domain, specifically $(-0.9,3.9) \times (-1.2,1.2)$. For each angle of attack, one base (undisturbed) case and a total of 20 gust cases are considered, generating a dataset with 5 base cases and 100 gust cases. For each case, 745 snapshots are uniformly sampled over time, resulting in a total of 78,225 data points. Each computation is performed over 15 convective times from the instant the disturbance is introduced to the flow.

\subsection{Low-order representation of flow} \label{latent_space}
\begin{figure*}[h!]
\centering
\includegraphics[width=1.0\textwidth]{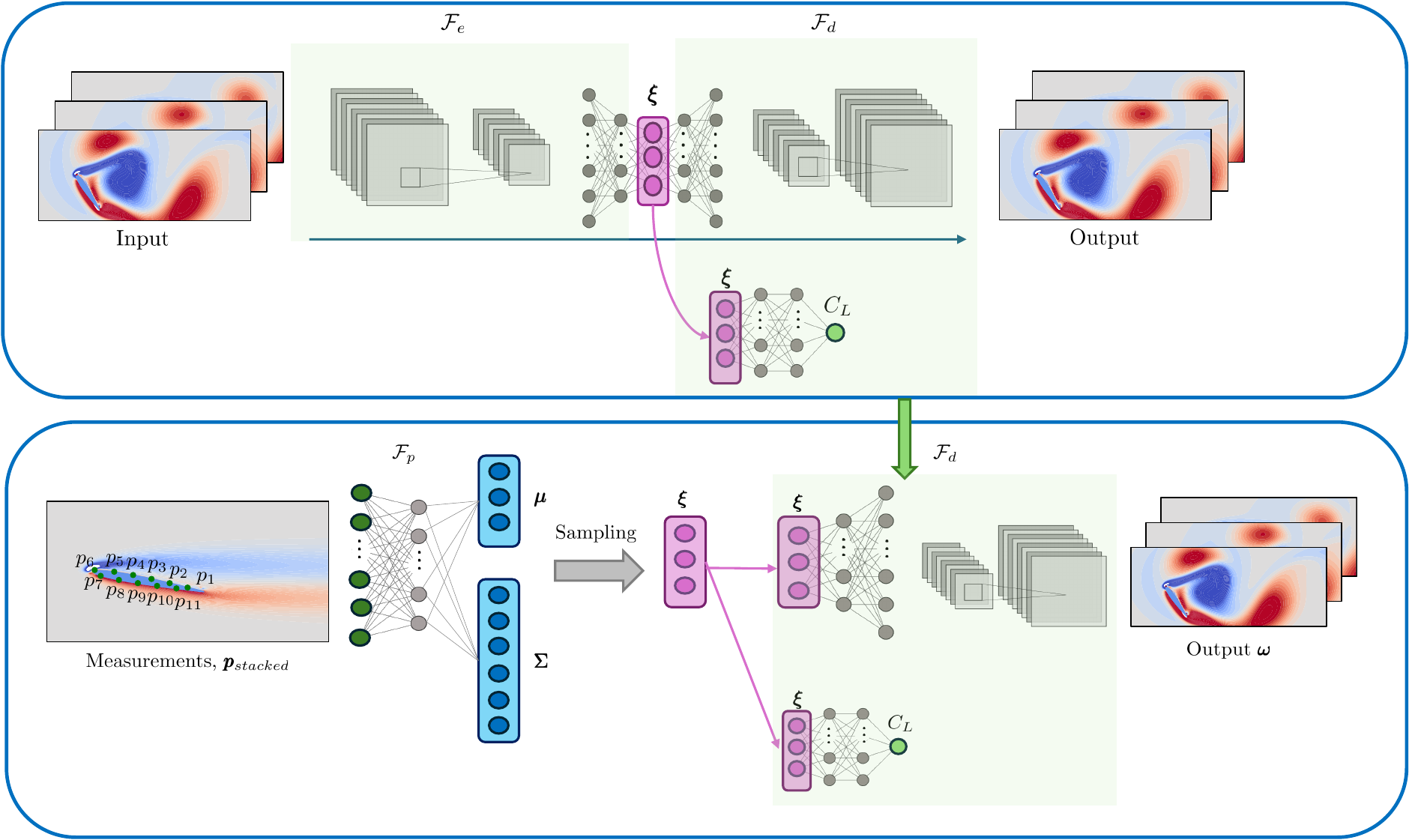}
\caption{\label{fig:network} Overview of the network architecture in the present study. The flow field data is compressed into a three-dimensional latent vector, denoted as $\bo{\xi}$, using the lift-augmented autoencoder. The architecture of this autoencoder is shown in the top panel. In the subsequent step, a pressure-based (MLP) network is trained to estimate the statistical parameters of a model distribution in the latent space, as illustrated on the left side of the bottom panel. This estimated latent vector sampled from the model distribution is then input into the decoder component of the autoencoder (top panel) to reconstruct both the vorticity field and the lift coefficient, as depicted on the right side of the bottom panel.}
\end{figure*}
Training a model to map sparse, low-dimensional measurements to a high-dimensional flow field necessitates deep architectures with numerous layers and nodes, which can lead to computational intractability. This challenge is particularly pronounced in the context of uncertainty quantification, as the cross-correlations within the high-dimensional output significantly increase the data size, complicating training and computational efficiency. Building on the findings of \citet{fukami2023grasping}, we utilize a non-linear lift-augmented autoencoder to derive a low-dimensional representation of the high-dimensional gust-encounter flow field. Before tackling uncertainty quantification, we first focus on data-driven flow compression to identify this low-dimensional space, which effectively captures the key physics of vortex-gust-airfoil interactions. We adopt the network architecture described in \citet{fukami2023grasping}, which integrates a Multi-Layer Perceptron (MLP) with a Convolutional Neural Network (CNN). As illustrated in the top panel of Fig.~\ref{fig:network}, our data compression framework comprises an encoder $\mathcal{F}_e$ that reduces the high-dimensional data to a low-dimensional latent vector $\bo{\xi} \in \mathbb{R}^l$ where $l\ll n$, with $n$ the data dimension. (In our applications in this paper, $l=3$, as in \citet{fukami2023grasping}; the appropriateness of this choice will be discussed below.) This is followed by two decoders $\mathcal{F}_d$: one for reconstructing the vorticity field $\bo{\omega}$ and another for estimating the lift coefficient $C_L$.

The weights are determined by solving an optimization problem that involves minimizing the loss function, defined as:
\begin{equation}\label{eq:weights}
\pmb{W} = \text{argmin}_{\bo{W}} \left( ||\bo{\omega} - \hat{\bo{\omega}}||_2 + \beta ||C_L - \hat{C}_L||_2 \right),
\end{equation}
where the hat over the parameters denotes the predicted field using the NN. Here, $\beta$ is a coefficient that balances the losses associated with vorticity and lift, with its value set to $0.05$ as specified in \citet{fukami2023grasping}. The weights are optimized using the Adam optimizer. The non-linear activation function employed is the hyperbolic tangent function.

Eighty percent of the data is allocated for training, while the remaining twenty percent is used for validation and testing. The training process utilizes \texttt{Early Stopping}, as implemented in the TensorFlow library, to halt training if the validation loss does not improve for 200 consecutive epochs.

\subsection{Flow reconstruction from sparse sensors}
The reduced-order latent vector extracted in the previous section is critical for effectively capturing the vorticity field and lift. This latent vector serves as a compact, informative representation of the complex, high-dimensional flow dynamics, enabling efficient analysis and prediction while still preserving the essential features of the flow dynamics in the network weights. Thus, we wish to learn how observable data can be used to estimate the latent state of the flow. As illustrated in Fig.~\ref{fig:network} left side of the bottom panel, we map surface pressure measurements to the latent space using a NN with MLP hidden layers. By leveraging the significant reduced dimensionality as the latent space representation, we can uniquely estimate the flow field and lift coefficient from available sensor data. 

An overview of the network architecture for estimation purposes is shown in the bottom panel of Fig.~\ref{fig:network}. The following sections will demonstrate that capturing heteroscedastic uncertainty—--characterized by noise-dependent variability—--requires the model to output the statistics of the low-order flow in the latent space, rather than simply providing a pointwise prediction of the latent state. In this study, we predict the mean and covariance matrix as the first two moments of a multivariate normal distribution in the latent space. To ensure the flow estimation is independent of the specific position of the airfoil in the domain and to avoid the estimator's confusion about the angle of attack, we incorporate the sensors $x$ and $y$ positions in a global reference frame along with the measurements and denote them by $\meas$. By stacking the sensor positions with the pressure measurements, we account for the relative positioning of the sensors. This integration enhances the network’s ability to generalize across different spatial configurations, making flow reconstruction robust to changes in the airfoil's location. In this study, 11 pressure measurements are utilized, resulting in an input vector of size 33 for the pressure-based estimator (the $x$ and $y$ sensor coordinates along with the readings themselves). Weight regularization is employed to constrain the magnitude of the network's weights, preventing them from becoming excessively large. This helps control the model's uncertainty, ensuring more stable and reliable predictions.

Once the stacked pressure measurements and sensor positions are mapped to the latent space, the pre-trained decoder component of the NN, denoted by $\mathcal{F}_d$, from the lift-augmented autoencoder as detailed in Section \ref{latent_space}, can be adopted to reconstruct the vorticity field and lift from this latent vector. This reconstruction process, depicted in Fig.~\ref{fig:network} on the right side of the bottom panel, leverages the network's ability to translate the reduced-order representation back into the high-dimensional flow field and aerodynamic forces with high fidelity. The advantage of this general approach is its capability to reconstruct the flow field and lift from limited measurements without sacrificing generality or requiring a large network. By focusing on the latent vector, we ensure that the essential characteristics of the flow are captured, enabling accurate predictions even with sparse data. 

\subsection{Informative directions of measurements} \label{dominant_direction}
Not all measurements contribute equally to estimating the flow field at each time step. The most responsive measurement subspace contains the most informative directions for capturing the flow field's dynamics. These directions reflect how the estimated flow field is most sensitive to particular weighted combinations of sensor measurements. This subspace evolves due to the transient nature of vortex shedding behind the airfoil and gust-airfoil interactions.

In this section, we conduct a sensitivity analysis of our NN model, which maps inputs $\vecx$ to outputs $\vecy$ through a non-linear function $\vecy=\f(\vecx;\bo{W})$, where $\bo{W}$ represents the network parameters, including weights and biases in a typical deep network. In this study, the inputs $\vecx \in \mathbb{R}^d$ correspond to the measurements augmented with the sensor coordinates, denoted by $\meas$, while the output $\vecy \in \mathbb{R}^l$ of the NN is the $l$-dimensional latent vector $\lat$. The non-linear mapping from inputs to outputs $\f : \mathbb{R}^d \rightarrow \mathbb{R}^l$ in this context refers to $\mathcal{F}_p$ described in the previous section, but with the output restricted to the mean prediction, i.e., we assume a deterministic form of the NN for this discussion. For clarity and consistency, we use $\vecx$, $\vecy$, and $\f$ to refer to inputs, outputs, and the mapping between them, respectively, in the derivations presented in this section.

For stochastic inputs and outputs influenced by sensor noise, the primary directions in the measurement space in which measurement fluctuations are most informative of variations in the flow field (via the latent space) can be identified using the Gramian matrix of the Jacobian in the input space, defined as follows \citep{quinton2024jacobian}:
\begin{equation} \label{eqcx1}
    \bo{C_x} = \mathbb{E} [\nabla \f(\vecx)^T \nabla \f(\vecx)],
\end{equation}
where $\mathbb{E}[\cdot]$ denotes the expectation with respect to the noisy input $\vecx$, and $\nabla \f$ represents the Jacobian matrix that describes the derivatives of the output vector $\vecy$ with respect to the input vector $\vecx$. We call $\bo{C_x}$ the \emph{measurement space Gramian}, and revisit Eq.~\eqref{eqcx1} to obtain
\begin{equation} \label{eqCx}
    \bo{C_x} = \int \nabla \f(\vecx)^T \nabla \f(\vecx) d\pi(\vecx),
\end{equation}
where $\pi(\vecx)$ is the probability density function of the input $\vecx$. The matrix $\bo{C_x}$ is positive semi-definite (PSD), and its eigendecomposition can be written as $\bo{C_x} = \bo{U} \bo{\Lambda}_x^2 \bo{U}^T$, where $\bo{U} \in \mathbb{R}^{d \times d}$ contains the eigenvectors with $\bo{\Lambda}_x^2$ the associated eigenvalues. The eigenvectors of the \emph{measurement space Gramian} identify the subspace spanned by the dominant (i.e., most informative) directions of the measurements. For convenience, we assume that the eigenvalues $\bo{\Lambda}_x^2$ are in decreasing order. In practice, we approximate these integrals using the Monte Carlo method to compute $\bo{C_x}$. The calculations of Jacobians $\nabla \f \in \mathbb{R}^{l \times d}$ for a NN denoted here by $\mathcal{F}_p$ can be performed using automatic differentiation.

For the low-order representations, only the first $r_x \leq d$ eigenmodes for the measurement space are retained, which corresponds to dominant modes $\bo{U}_r$. The rank $r_x$ is tuned based on the decay of $\bo{\Lambda}_x^2$. Typically these ranks are set to achieve a threshold $\gamma \in [0,1]$ for the cumulative normalized energy of the eigenvalue spectra. The first $r_x$ eigenmodes of $\bo{C_x}$ correspond to the directions in the input space that most influence the prediction of the latent space. The components of an eigenmode represent the weights on the sensors' contributions to this mode; a larger magnitude component indicates a greater relative role for that sensor in detecting a disturbance.

\subsection{Quantifying aleatoric and epistemic uncertainties}
\label{sec:quantify-uncertainty}
In real-world applications, measurement noise is a common occurrence, and the pressure measurements in this study are no exception. This introduces uncertainty into the predictions, which is particularly critical for risk management applications where accurate quantification of uncertainty is essential. To assess the uncertainty in the outputs due to measurement noise, we define the conditional output distribution given the input and the network parameters $\pmb{W}$ as $\pi_a(\vecy|\vecx, \pmb{W})$, where the subscript $a$ denotes aleatoric uncertainty. Here, the noisy inputs are represented as $\vecx=\bar{\vecx}+\bo{\eta}$, with $\bar{\vecx}$ referring to clean data and $\bo{\eta}$ representing the random sensor noise. 

Aleatoric (or data) uncertainty, originating from inherent noise in the data, must be explicitly captured to enhance the network’s resilience to noisy inputs. In real-world scenarios, the probability distribution of outputs typically varies as a function of the inputs. This data-dependent nature of aleatoric uncertainty can be effectively addressed using heteroscedastic models, which incorporate learned loss attenuation to model input-dependent noise levels \citep{kendall2017uncertainties}. Consequently, the network architecture depicted in Fig.~\ref{fig:network} bottom panel is designed specifically for uncertainty quantification. It learns the parameters of a multivariate normal distribution, i.e. the mean denoted as $\bo{\mu} \in \mathbb{R}^l$, and the covariance matrix represented by $\bo{\Sigma} \in \mathbb{R}^{l \times l}$. During training, the network minimizes a heteroscedastic loss function based on the negative log-likelihood of a multivariate normal distribution:
\begin{equation} \label{eq:loss}
    \mathcal{L}_{dropout} = -\log \left(\mathcal{N} (\bo{y} | \hat{\bo{\mu}} , \hat{\bo{\Sigma}}) \right),
\end{equation}
where $\bo{y}$ represents the true latent vector, $\hat{\bo{\mu}}$ and $\hat{\bo{\Sigma}}$ are the predicted mean and covariance matrix, respectively. The multivariate normal distribution is defined as usual by
\begin{equation}
    \mathcal{N} (\bo{y} | \hat{\bo{\mu}} , \hat{\bo{\Sigma}})) = \frac{1}{\sqrt{(2\pi)^l \det (\hat{\bo{\Sigma}})}} \exp \left( -\frac{1}{2} (\bo{y} - \hat{\bo{\mu}})^T \hat{\bo{\Sigma}}^{-1} (\bo{y} - \hat{\bo{\mu}}) \right).
\end{equation}

It should be noted that directly predicting a covariance matrix poses challenges since the network might not inherently ensure that the matrix is symmetric and positive definite. To address this, we reformulate the prediction: instead of predicting the full covariance matrix, the network learns $l \times (l+1)/2$ elements of a lower-triangular matrix, $\bo{L}$, in the form of
\begin{equation}
    \bo{L} = 
    \begin{bmatrix}
        L_{11} & 0 & 0 & \cdots & 0 \\
        L_{12} & L_{22} & 0 & \cdots & 0 \\
        \vdots & \vdots & \vdots & \vdots & \vdots \\
        L_{1l} &  L_{2l} & L_{3l} & \cdots & L_{ll} \\
    \end{bmatrix}_{l \times l}.
\end{equation}
Considering that the dimension of the latent vector is denoted by $l$, the predictions are in the space $[\hat{\bo{\mu}}, \hat{\bo{L}}] \in \mathbb{R}^{l+l(l+1)/2}$. This guarantees uniqueness, symmetry, and positive-definiteness by constructing the covariance matrix as $\bo{\Sigma} = \bo{L} \bo{L}^T$. For numerical stability, the network predicts the logarithms of the squared diagonal elements, ensuring they remain positive, and directly outputs the off-diagonal elements. By constructing the covariance matrix this way, we ensure it remains positive definite and approximates the Cholesky decomposition. Additionally, to improve the network's robustness to input noise, we employ data augmentation during training, injecting Gaussian random noise into the inputs to simulate real-world conditions and help the model better capture aleatoric uncertainty.

Traditional deep learning models provide point-estimate predictions with overconfidence, and do not typically account for the uncertainty in the fitted model, called epistemic (or model) uncertainty. The main goal in model uncertainty is finding the conditional distribution over the parameters $\bo{W}$ of the NN for a given dataset of inputs $\bo{X}$ and outputs $\bo{Y}$, i.e., $\pi(\bo{W}|\bo{X},\bo{Y})$. Among the different ways to estimate uncertainty in the NN model, the Bayesian paradigm provides a powerful mathematical framework. Indeed, Bayes' rule expresses the desired conditional distribution as a posterior distribution, starting from a prior $\pi(\bo{W})$ over the weights: 
\begin{equation}
    \pi(\bo{W}|\bo{X},\bo{Y}) = \frac{\pi(\bo{X},\bo{Y}|\bo{W}) \pi(\bo{W})}{\pi(\bo{X},\bo{Y})}.
\end{equation}
Here, $\pi(\bo{X},\bo{Y}|\bo{W})$ is the conditional probability of the data, given a particular set of weights. The denominator in this equation is the marginal distribution over the space of model parameters $\bo{W}$. Calculating this marginal distribution is challenging. Basically, there are two primary approaches to address this difficulty: Markov Chain Monte Carlo (MCMC), which samples from the true posterior and avoids the need for the denominator by only relying on comparison; and Variational Inference (VI) \citep{blei2017variational}, which approximates the posterior with a known family of distributions denoted by $q(\bo{W})$. MCMC converges very slowly in large and complex networks and requires a large number of samples to achieve convergence. As an efficient alternative for obtaining the posterior distribution, deep neural networks with dropout applied before every weight layer have been shown to be mathematically equivalent to approximate VI in a deep Gaussian process (GP) \citep{gal2016dropoutbayesianapproximationappendix}. This procedure, known as MC dropout, uses a variational distribution defined for each weight matrix as follows:
\begin{equation}
\begin{split}
    z_{i,j} &\sim \text{Bernoulli} (p_i), \\
    \bo{W}_i &= \bo{M}_i \cdot \text{diag} (\bo{z}_i),
\end{split}
\end{equation}
with $z_{i,j}$ referring to the random activation coefficient for the
$j$-th neuron in the
$i$-th layer (1 with probability $p_i$ for layer $i$ and 0 with probability ($1-p_i$)), and $\bo{z}_i$ being the random activation coefficient vector, containing all $z_{i,j}$ for layer $i$. The matrix $\bo{M}_i$ is the matrix of weights before dropout is applied. This approximate distribution, as proven in \citet{gal2016dropoutbayesianapproximationappendix}, minimizes the Kullback-Leibler divergence ($D_{KL}$), which measures the similarity between two distributions $D_{KL}(q(\bo{W})||\pi(\bo{W}|\bo{X},\bo{Y}))$.

In MC dropout, a subset of activations is randomly set to zero during training, and the same values are used in the backward pass to propagate the derivatives to the parameters. In typical NNs, dropout is usually turned off during evaluation. However, leaving it on during inference produces a distribution for the output predictions, allowing for the estimation of uncertainty in the predictions in the form of 
\begin{equation}\label{eq:marginal}
    \pi_e(\vecy|\vecx,\datax,\datay) = \int \pi(\vecy|\vecx,\bo{W}) q(\bo{W}) d\bo{W}.
\end{equation}
The subscript $e$ in $\pi_e(\vecy|\vecx,\datax,\datay)$ denotes epistemic uncertainty. Using the Monte Carlo method, multiple stochastic forward passes are performed to approximate this integral, effectively sampling from the posterior distribution. 

Though MC dropout is utilized to quantify model uncertainty, the dropout layers in the network shown in the bottom panel of Fig.~\ref{fig:network} remain active during both training and inference, and thus affect both forms of uncertainty quantification. As described earlier, aleatoric uncertainty is quantified with a network (denoted by $\mathcal{F}_p$) that produces two outputs: the mean of the latent vector, $\bo{\mu} \in \mathbb{R}^l$, and the covariance matrix of the latent vector $\bo{\Sigma} \in \mathbb{R}^{l\times l}$. By applying different instances of MC dropout, $\mathcal{F}_p$ produces a distribution of the output, and this can be assumed to be multivariate Gaussian with its statistics represented by the expected value and covariance of the output samples obtained from $T$ stochastic forward passes. As such, during inference, the aleatoric predictive distribution, marginalized over the network weights, is measured by
\begin{equation} \label{eq:pi_a}
    \pi_a(\vecy|\vecx,\datax,\datay) = \mathcal{N} \left(\vecy;\frac{1}{T} \sum_{k=1}^T \hat{\bo{\mu}}_k, \frac{1}{T} \sum_{k=1}^T \hat{\bo{\Sigma}}_k \right).
\end{equation}
Additionally, the epistemic predictive distribution is computed by
\begin{equation} \label{eq:pi_e}
    \pi_e(\vecy|\vecx,\datax,\datay) = \mathcal{N} \left(\vecy;\frac{1}{T} \sum_{k=1}^T \hat{\bo{\mu}}_k, \text{Cov} \left( \{\hat{\bo{\mu}}_k \}_{k=1}^T \right) \right),
\end{equation}
with $\{ \hat{\bo{\mu}}_k, \hat{\bo{\Sigma}}_k \}_{k=1}^T$ a set of $T$ sampled outputs. We emphasize that the output covariance associated with the aleatoric uncertainty is predicted directly by the network and averaged over the dropout passes (the mean of the covariances), while the output covariance of the epistemic distribution follows from the spread in the pointwise predictions of the output over these passes (the covariance of the means). 

To quantify the uncertainty in the output most influenced by variations in the input during inference, we introduce noise $\bo{\eta}$ aligned with the principal directions of measurement variation. These directions are identified by the matrix $\bo{U}_r$, which contains the eigenvectors associated with the largest eigenvalues of the measurement space Gramian, as discussed in Section \ref{dominant_direction}. The rank $r_x$ is determined based on capturing $99\%$ of the cumulative energy spectrum of the eigenvalues $\bo{\Lambda}_x^2$ of the measurement space Gramian $\bo{C_x}$. Typically, the first two eigenvalues were observed to account for over $99\%$ of the energy, and often the first eigenvalue alone was sufficient. Thus, the input noise is modeled as $\bo{\eta} \sim \zeta \bo{U}_r$, where $\bo{U}_r$ represents the dominant modes of the measurements, and $\zeta \sim \mathcal{N}(0,\sigma_x^2)$ is a random coefficient with $\sigma_x^2$ representing the variance in the sensor noise.

After training the network with corrupted sensor data, as described earlier, the distributions of the latent vector are computed using Eqs.~\eqref{eq:pi_a} and \eqref{eq:pi_e}. From these distributions, $M$ samples of latent vectors $\hat{\bo{\xi}}_i$ are drawn and passed through the decoder $\mathcal{F}_d$ (see Fig.~\ref{fig:network}, bottom panel) to reconstruct the corresponding vorticity and lift samples, $\{ \hat{\bo{\omega}}_i, \hat{C}_{L,i} \}_{i=1}^M$. The reconstruction procedure is defined as
\begin{equation}
\begin{aligned}
    \hat{\bo{\xi}}_i &\sim \pi_u(\vecy|\vecx,\datax,\datay) \qquad \text{for} \ i = 1, 2, \cdots, M \\ 
    \{ \hat{\bo{\omega}}_i, \hat{C}_{L,i} \} &= \mathcal{F}_d(\hat{\bo{\xi}}_i),
\end{aligned}
\label{eq:uncertainty-reconstruct}
\end{equation}
where the subscript $u$ can be either $a$ for aleatoric or $e$ for epistemic. We will assume that the statistics of the reconstructed vorticity and lift follow a normal distribution, described by the mean and variance of the reconstructed samples. In the case of vorticity, this normal distribution is local to each grid point (pixel). To quantify the performance of this reconstruction in either type of uncertainty quantification scenario, we will compute the log-likelihood of the true vorticity at each pixel and average over the pixels; large values of this averaged log-likelihood indicate two qualities: that the overall uncertainty is small and that the true vorticity falls within the uncertainty bounds. As a result, aleatoric and epistemic uncertainties can be quantified independently for any variable of interest, including the latent vector, lift force, and vorticity field. This distinction allows for a deeper understanding of how input noise and model uncertainty affect different aspects of the reconstructed flow fields. 

In the sensor-based prediction performed by the network $\mathcal{F}_p$, the \verb|ReLU| activation function is used for its ability to introduce non-linearity while effectively avoiding the restricted uncertainty range often associated with \verb|TanH| activation. Again, eighty percent of the data is allocated for training, while the remaining twenty percent is used for validation and testing. To optimize training and prevent overfitting, \texttt{Early Stopping} is incorporated to stop training if there is no improvement in validation loss for 500 consecutive epochs. After experimentation, a regularization constant of $10^{-7}$ and a dropout rate of $0.05$ were identified as optimal, yielding the highest average likelihood during training.

\section{Results}\label{results}
Viscous flow over a NACA 0012 airfoil is simulated using the immersed layers method proposed by \citet{eldredge2022method}, both in the presence and absence of disturbances. Detailed descriptions of the flow solver and data generation process are provided in Section \ref{probstatement}. The simulation covers 105 cases, resulting in a dataset of 78,225 points, with 3,725 points corresponding to base cases and the remainder to random disturbed flow cases. For estimation using deep learning, we collect the vorticity field $\bo{\omega}$, lift coefficient $C_L$, pressure coefficient $C_p$, and the coordinates of surface pressure sensors $(x_{sens}, y_{sens})$. We deploy 11 evenly-spaced sensors on both sides of the airfoil, as shown in Fig.~\ref{fig:gust_approach}. Stacked with their locations, the input measurement vector $\meas$ has a dimension of $33$.

To gain a deep understanding of sensor response and accurately identify the regions most affected by gust interactions, a more detailed analysis is necessary. This detailed assessment will be addressed in subsequent discussions within this section.

\begin{figure*}[tb!]
\includegraphics[width=1.0\textwidth]{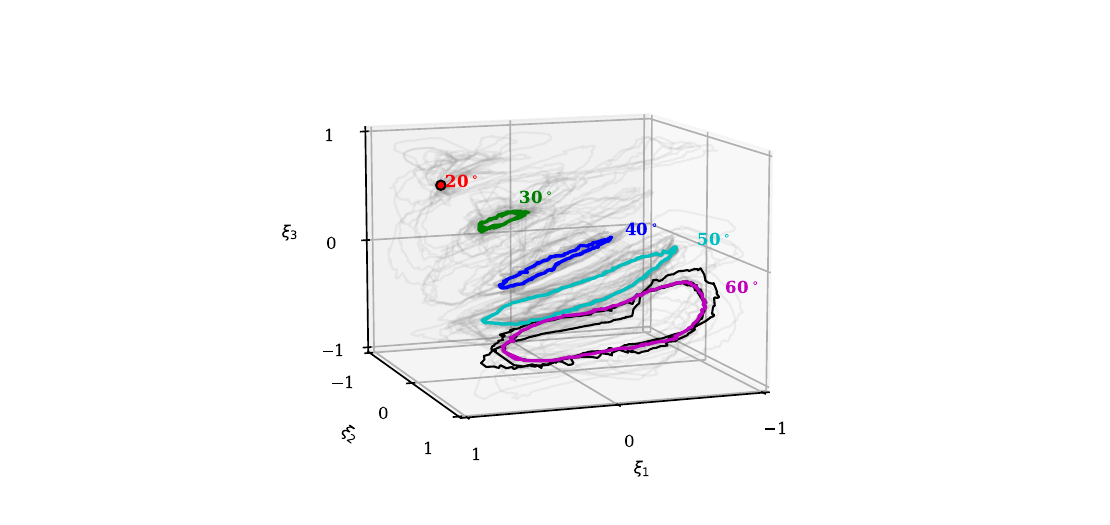}
\caption{\label{fig:x_lat_3d} Low-order representation of flow data is presented with undisturbed cases highlighted in color for five angles of attack. The light gray paths indicate disturbed cases with the black path highlighting one of them.}
\end{figure*}

\subsection{Extracting low-order representation of flow}\label{latent_variables}
To manage computational expenses associated with uncertainty quantification, we initially reduce the dimensionality of the collected flow data via the lift-augmented autoencoder. The results are presented in Fig.~\ref{fig:x_lat_3d}. It displays the projected flow field in a three-dimensional latent space, illustrating the discrete pathlines for each case. This projection demonstrates the distinguishability of different cases in 3D space. Notably, the coordinate $\xi_3$ correlates with the airfoil's angle of attack, while the other two coordinates reflect the vortical variations in the flow. The limit-cycle behavior is evident in the trajectories, particularly in the gust cases. In these cases, the pathline deviates from the undisturbed trajectory as the gust passes over the airfoil and eventually returns to the periodic undisturbed orbit once the gust leaves the domain. This dynamic is clearly illustrated for an airfoil encountering a gust at $\alpha=60^\circ$, as shown by the black trajectory in Fig.~\ref{fig:x_lat_3d}.

\subsection{Sensor-based flow reconstruction}
Stacked with their coordinates $(x, y)$, the sensor readings—--denoted by $\meas$ as a 33-dimensional vector—--are mapped to the three-dimensional latent space $\lat$, as illustrated on the bottom-left side in Fig.~\ref{fig:network}. These latent variables, extracted in Section \ref{latent_variables}, correspond to the compressed flow field and lift. Given that both the inputs and outputs are vectors, we employ a MLP network to model the mapping, denoted as $\mathcal{F}_p$. To quantify the network's uncertainty in its predictions, we utilize MC dropout, incorporating a dropout layer after each dense layer in the MLP network. During both training and inference, dropout layers are active to model epistemic uncertainty. Furthermore, to account for uncertainty in the input measurements, the network is trained to predict the covariance matrix in the latent space as well. The details of the procedure are described in Section~\ref{sec:quantify-uncertainty}, and the details of the network architecture itself are provided in Table~\ref{tab:network_architecture} in the appendix.

In the context of MC dropout, the dropout rate $(1-p)$ is a hyperparameter that is optimized to maximize the log-likelihood (or equivalently, minimize the loss given in Eq.~\eqref{eq:loss}). Interestingly, it was observed that the dropout rate has minimal impact on the log-likelihood. Consequently, a fixed dropout rate of $0.05$ was selected for training the MLP network $\mathcal{F}_p$.

\subsubsection{Sensitivity analysis to flow disturbances}
To analyze how sensor variations respond to disturbances in the flow structures, we identify the most informative direction within the measurement space Gramian, $\bo{C_x}$, as defined in Eq.~\eqref{eqCx}. In this subsection, we aim to identify the dominant eigenvectors within the measurement space. To achieve this, we utilize the same network architecture described in Table \ref{tab:network_architecture}, but we modify the final layer to directly output the latent variables deterministically (i.e., we omit the prediction of covariance). This network is trained on clean data using a mean squared error (MSE) loss function to optimize the model weights effectively. During the evaluation phase, dropout is disabled to ensure consistent predictions with the same inputs. Additionally, we approximate the integral in Eq.~\eqref{eqCx} through Monte Carlo sampling, employing 100 samples of noisy measurements during inference to obtain a robust estimate of the dominant eigenvectors. The measurement noise is modeled as independent and identically distributed (i.i.d) white noise with a mean of zero and a variance of $2.5 \times 10^{-5}$, corresponding to a measurement accuracy of $0.15\%$ for the maximum pressure reading. The sensor coordinates on the airfoil are assumed to be accurately measured (i.e., they are assigned zero variance). Our analysis reveals that the first two eigenmodes of the Gramian account for more than $99 \%$ of the cumulative energy spectrum of the eigenvalues. The first eigenmode indicates the direction in the measurement space that is most informative for latent vector estimation.

\begin{figure*}[tb!]
\includegraphics[width=1.0\textwidth, trim=0 30 0 0, clip]{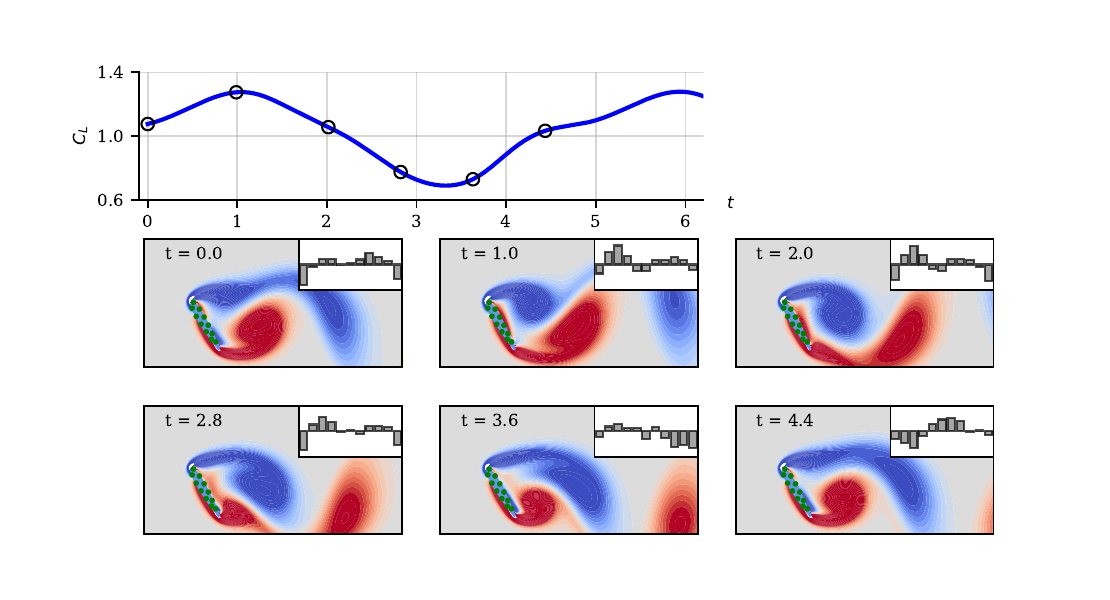}
\caption{\label{fig:sensor_importance_base_60} Periodic variation of lift and the first mode of surface pressure measurements over time for undisturbed flow at an angle of attack of $\alpha=60^\circ$. The bar plot corresponds to the first eigenmode of the measurement space Gramian; each bar represents a sensor, with their order corresponding to the numbering scheme depicted in Fig.~\ref{fig:gust_approach}, arranged sequentially from left to right.}
\end{figure*}

We first examine the undisturbed (base) cases for flow over an airfoil. The importance of pressure sensors in detecting flow structures around the airfoil, and the amount of information they convey, is illustrated in Fig.~\ref{fig:sensor_importance_base_60} for an angle of attack of $\alpha=60^\circ$, exemplifying an unsteady vortex shedding case. The figure illustrates the placement of pressure sensors on the airfoil, marked by green circles. Accompanying each snapshot is a bar chart that highlights the relative importance of the pressure measurements in estimating the latent vector. This significance is derived from the first eigenmode of the Gramian matrix in the measurement space. Each bar represents a sensor, with their order corresponding to the numbering scheme depicted in Fig.~\ref{fig:gust_approach}: the first bar represents the sensor on the upper surface nearest the trailing edge, and the subsequent bars correspond to sensors that proceed counter-clockwise around the airfoil, culminating in the sensor closest to the trailing edge on the lower surface. The height of the bar, whether positive or negative, indicates its relative importance to estimating the latent state.

According to Fig.~\ref{fig:sensor_importance_base_60}, in the interval $t \in [0,2.8]$ a clockwise leading-edge vortex (LEV) grows, sustained by the shear layer from the leading edge. As the vortex grows to extend across the entire chord, the airfoil experiences its highest lift from $t = 1$ through 2. During this interval, the sensors 2 through 4 near mid-chord of the suction side become most impactful. This interval also corresponds to the generation and extension of counter-clockwise secondary vorticity along the entire suction side. After $t = 2$, the LEV begins to shed from the airfoil, and the importance of the suction-side sensors begins to diminish in favor of those on either side nearest the trailing edge (1 and 11), where a new trailing-edge vortex (TEV) begins to emerge. As the TEV develops from $t = 2.8$ through 3.6 and lift is at its lowest level, the sensors 9 through 11 on the pressure side become important. At $t = 4.4$, new secondary vorticity develops under the TEV and a new vortex develops at the leading edge, and the sensors in the respective regions (1-3 and 5-8) are most informative. The cycle returns to the state at $t = 0$, when the trailing-edge sensors (1 and 11) become most important.

\begin{figure*}[bt!]
\includegraphics[width=1.0\textwidth, trim=25 50 25 50, clip]{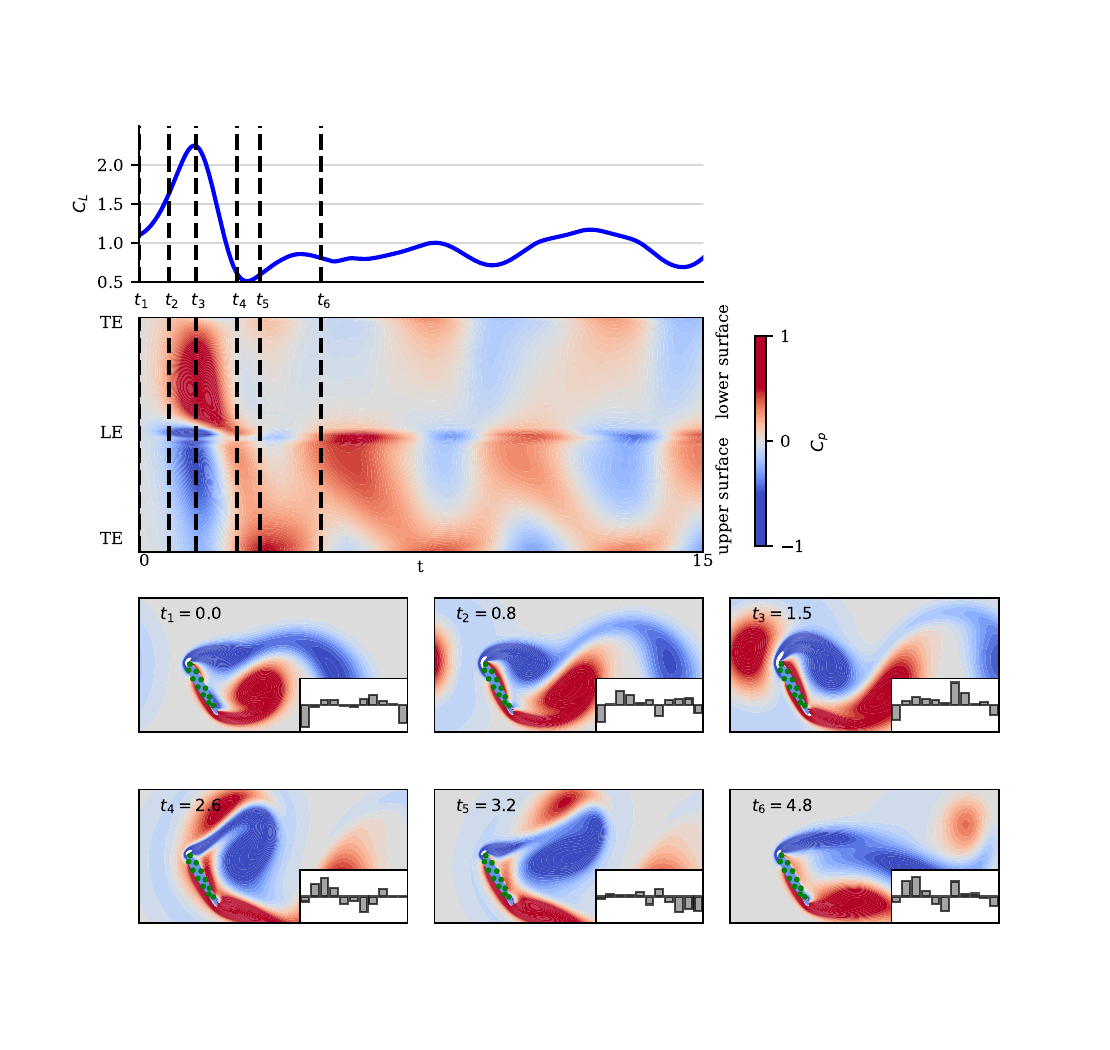}
\caption{\label{fig:sensor_importance_positive_gust_60} The figure illustrates the primary mode of pressure measurements at six different snapshots as a positive gust interacts with the airfoil. The first panel depicts the temporal variation in lift. The spatiotemporal map of the pressure coefficient in the second panel, with the base flow subtracted, provides insight into how sensors respond to gust-airfoil interactions. The final two panels present vorticity contours alongside sensor placements. Each bar represents a sensor indicating its value in the dominant eigenmode, with their order corresponding to the numbering scheme depicted in Fig.~\ref{fig:gust_approach}, arranged sequentially from left to right. The conditions are an angle of attack of $\alpha=60^\circ$, and gust characteristics of ($G=0.9$, $2R/c=0.98$, $y_o/c=-0.06$).}
\end{figure*}

In contrast to cases at large angles of attack, in which the first measurement mode varies periodically over time, at steady flow conditions---such as those for an airfoil at an angle of attack of $\alpha=20^\circ$ or less---we find (though omit the plots for brevity) that sensors closest to the trailing edge on either side (1 and 11) are most informative.

\begin{figure*}[bt!]
\includegraphics[width=1.0\textwidth, trim=25 50 25 50, clip]{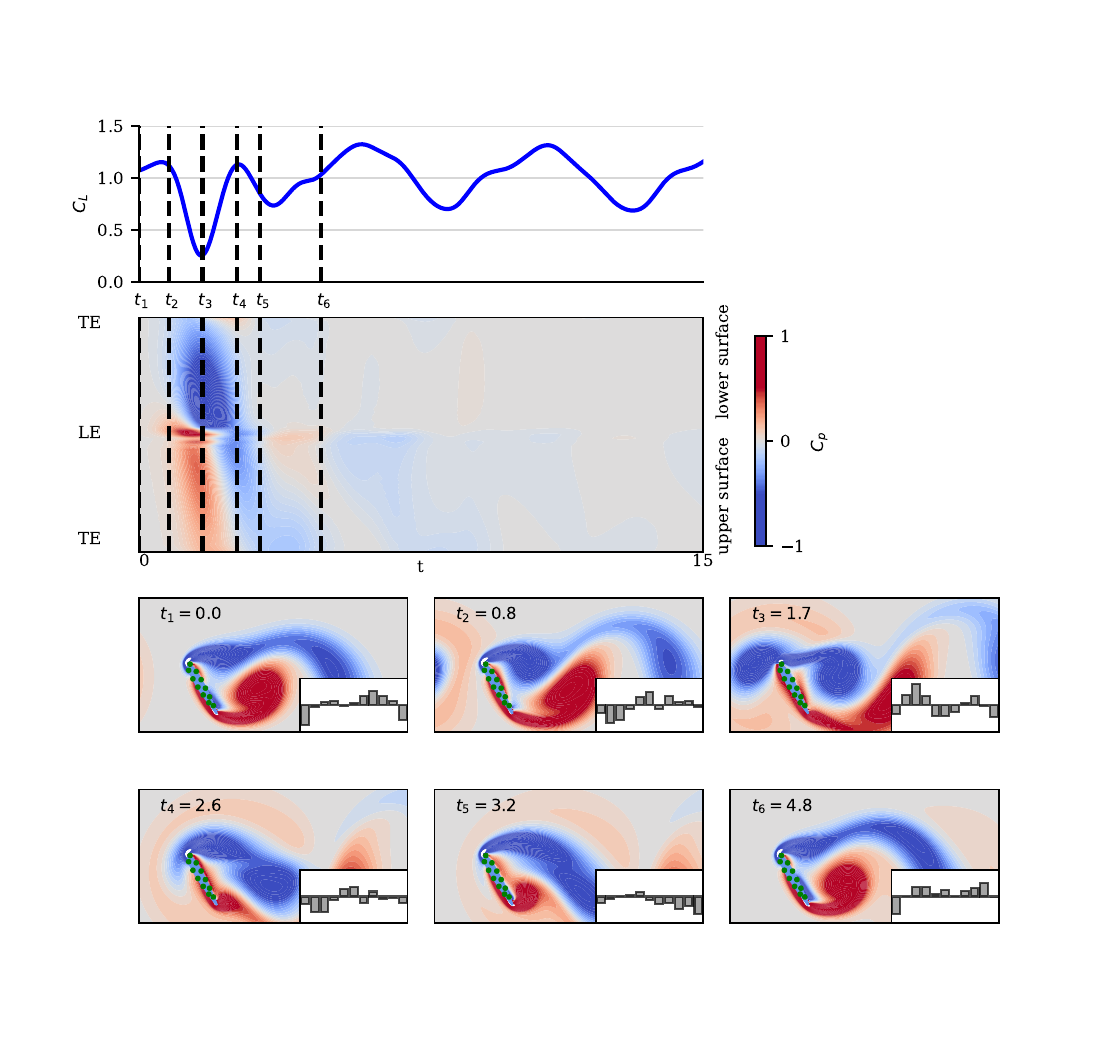}
\caption{\label{fig:sensor_importance_negative_gust_60} The figure illustrates the primary mode of pressure measurements at six different time snapshots as a negative gust interacts with the airfoil. The first panel depicts the temporal variation in lift. The spatiotemporal map of the pressure coefficient in the second panel, with the base flow subtracted, provides insight into how sensors respond to gust-airfoil interactions. The final two panels present vorticity contours alongside sensor placements. Each bar represents a sensor indicating its value in the dominant eigenmode, with their order corresponding to the numbering scheme depicted in Fig.~\ref{fig:gust_approach}, arranged sequentially from left to right. The conditions are an angle of attack of $\alpha=60^\circ$, and gust characteristics of ($G=-0.98$, $2R/c=0.77$, $y_o/c=-0.26$).}
\end{figure*}

The sensitivity of sensors to changes in the vortical structures becomes significantly more complex and transient during gust encounters over an airfoil. The top panel in Figure~\ref{fig:sensor_importance_positive_gust_60} illustrates the evolution of lift over time for a counterclockwise (positive) disturbance. In this scenario, there is an initial increase in lift as the gust approaches the airfoil, occurring around $t_3=1.5$. The lift force reaches its lowest point when the center of the vortical structure aligns with the trailing edge of the airfoil, at an instant between $t_4=2.3$ and $t_5=3.6$. This is then followed by an increase in lift due to tail-induced effects. For a clockwise (negative) gust, the sequence of events and their impact on the lift is reversed, as shown in Fig.~\ref{fig:sensor_importance_negative_gust_60} top panel. 

The central panel in Figure~\ref{fig:sensor_importance_positive_gust_60} illustrates a spatiotemporal distribution of the pressure coefficient for an airfoil subjected to a positive gust, with the baseline flow pressure subtracted for enhanced clarity. This visualization highlights distinct high-pressure regions on the pressure side and low-pressure zones on the suction side of the airfoil. A striking observation emerges: the positive, counterclockwise gust disturbance distinctly imprints on the pressure field, elevating pressure on the lower surface while diminishing it on the upper surface. This differential effect creates a stronger upward pressure gradient, ultimately enhancing the lift. These findings reinforce the rationale behind our strategic placement of pressure sensors to effectively capture flow perturbations around the airfoil. The strong interaction between the gust and the LEV disrupts the typical vortex dynamics, delaying the shedding process. This disruption is evident in the periodic non-zero pressure regions observed on the map, which persist even after the gust has exited the domain. Such behavior reflects a significant shift in the periodic vortex shedding pattern induced by the gust, underscoring the lasting impact of gust-driven disturbances on the flow field.

To explain some of the features we observe in the dominant eigenmodes, it is important to emphasize that the informativeness of pressure sensors as assessed by the Gramian is intrinsically tied to variations in their readings. These readings capture the evolving flow structure represented in the reduced-dimensional latent space. Consequently, the difference in the pressures across successive snapshots reflects the information conveyed in estimating flow states. Observing the vorticity fields alongside the sensor importance bars in the bottom two panels of Figure~\ref{fig:sensor_importance_positive_gust_60}, we can trace the impact of gust passage on pressure readings. When the gust tail reaches the leading edge at $t_2 = 0.8$, nearly all sensors—--particularly those near the leading edge—--register the flow changes. However, the tail's influence is weaker compared to the dominant effects of the primary LEV and TEV. Therefore, the sensors with the highest eigenmode response remain predominantly influenced by the primary vortices, similar to the base flow case. As the gust core interacts with the airfoil and the LEV/TEV beyond $t_2$, the dominant eigenmode of the pressure deviates significantly from the undisturbed scenario. At $t_3 = 1.5$, the pressure distribution plot reveals the steepest change in time on the lower side near the leading edge, making sensors 7 and 8 the most responsive to local flow changes. 
The interaction between the positive gust and the LEV leads to the formation of a vortex pair with unequal strengths, which moves downstream with the flow. At this time until the gust core moves approximately one chord length away at approximately $t_4=2.6$, sensors on the suction side become more informative due to substantial flow changes occurring there. When the gust core drifts further from the airfoil, sensors on the pressure side regain prominence, as seen at $t_5=3.2$. Eventually, the sensor responses transition back to the periodic behavior typical of the undisturbed flow once the gust exits the domain.
During this recovery phase, the pressure distribution plot reveals a pronounced change in time on the suction surface. This is attributed to the substantial adjustments in the primary edge vortices and boundary layer on this side as the flow reverts to its baseline state. At $t_6=4.8$, sensors near both the trailing and leading edges exhibit heightened activity, reflecting their critical role in capturing the dominant flow changes during this transitional phase.

The evolution of the first dominant mode of the measurement space Gramian for a negative gust is in many ways similar to that of a positive gust discussed earlier. Figure~\ref{fig:sensor_importance_negative_gust_60} illustrates the variations in the pressure coefficient during the passage of the negative gust. The spatiotemporal map, which depicts the pressure coefficient with the baseline flow subtracted, reveals dynamics that are qualitatively the inverse of a positive gust. Notably, however, the gust-vortex interactions induced by the negative gust are less disruptive overall. 
As the negative gust approaches the airfoil, it decreases pressure on the lower surface while increasing pressure on the upper surface during the period between $t_2=0.8$ and $t_3=1.7$. This reversal in pressure distribution creates a downward-directed pressure difference across the airfoil, ultimately reducing the lift generated by the airfoil during this interval. This distinct interaction underscores the impact of gust polarity on aerodynamic performance. 
When the negative gust interacts with the LEV, the vortices begin to rotate around a shared center and eventually coalesce into a larger vortex. This merging process initiates around $t_3=1.7$ and concludes by $t_5=3.2$. The resulting larger vortex intensifies the strength of the LEV, significantly affecting the pressure distribution on the suction side. Consequently, sensors 1 through 6, located on the upper surface of the airfoil, exhibit increased sensitivity to the localized flow changes up to $t_5=3.2$. As the gust core moves beyond one chord length from the airfoil, the influence on the suction side diminishes, and sensors on the pressure side regain prominence. This shift is particularly evident at $t_5$, further marking the transition back to baseline conditions as the gust recedes from the domain.
In contrast to the positive gust scenario, a negative gust does not cause a lasting delay in vortex shedding. This is apparent in the spatiotemporal map of the pressure coefficient, which reverts to baseline levels after the negative gust leaves the domain.

This general trend has also been observed for the airfoil at the other four angles of attack. Overall, throughout the gust-airfoil-wake interaction, nearly all sensors play a role in capturing the reduced-order flow dynamics, with their contributions varying over time based on the evolving flow structures and gust effects. This dynamic redistribution of sensor importance underscores the complexity and time-dependent nature of gust-induced disturbances.

\begin{figure*}[bt!]
\includegraphics[width=1.0\textwidth]{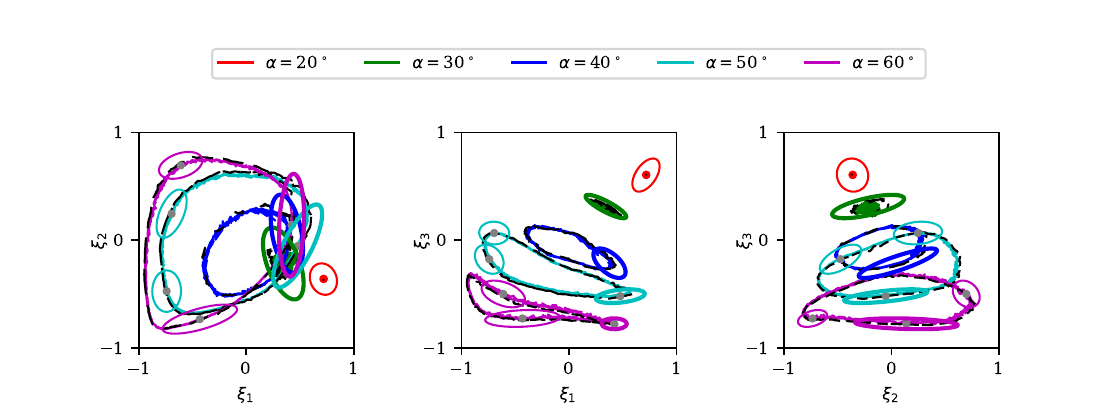}
\caption{\label{fig:aleatoric_uncertainty_x_lat_max_std} Predicted mean with $95 \%$ confidence ellipses of latent variables at a couple of instants for five undisturbed cases are shown. The solid-colored curves represent the mean of $\hat{\bo{\mu}}$, while the dashed black curves indicate the true trajectories extracted from the lift-augmented autoencoder. Thicker ellipses correspond to periods of maximum uncertainty. This figure showcases aleatoric uncertainty due to inherent noise in the input measurements.}
\end{figure*}

\subsubsection{Aleatoric Uncertainty}
As described in Section~\ref{sec:quantify-uncertainty}, to account for measurement noise in the reported predictions, we utilize the network trained with a heteroscedastic loss and active dropout, which maps surface pressure measurements to the mean and covariance matrix in the latent space representing the flow field and lift coefficient. During inference, the pressure coefficient, obtained from a high-fidelity numerical solver, is perturbed randomly along the most informative directions of the input measurements, denoted as $\bo{U}_r$, as identified in Section \ref{dominant_direction}; at most, there are only $r_x=2$ modes used. The noise coefficient $\zeta$ is modeled as an i.i.d white noise with zero mean and variance of $2.5 \times 10^{-5}$. We generate noisy inputs accordingly and perform $T=100$ forward passes through $\mathcal{F}_p$ to obtain a distribution of the mean and covariance of the latent variables. The expected value of the inferred mean over 100 samples is referred to as the predicted mean and represents the expected latent state, while the expected value of the covariance quantifies the aleatoric uncertainty of this state (see Eq.~\eqref{eq:pi_a}).

Figure~\ref{fig:aleatoric_uncertainty_x_lat_max_std} illustrates the trajectories of the predicted means of the latent variables alongside their $95 \%$ confidence ellipses at a small number of time instants, depicted in three plots representing the three coordinate planes of the latent space. These ellipses visualize the uncertainty distribution across five undisturbed cases. To compute these uncertainty bounds, we first perform a singular value decomposition (SVD) on the covariance matrix of the latent variables and then project the ellipsoid on the corresponding coordinate planes. The SVD extracts the principal axes of uncertainty, enabling us to align the ellipses along these dominant directions. The figure also includes the corresponding true trajectories extracted from the autoencoder, shown as black dashed lines for comparison. The figure clearly demonstrates that, despite the noise in the input measurements, the trained NN effectively maintains its robustness in predicting the low-dimensional representation of the flow from these measurements. The uncertainty ellipses further provide more information about the effect of measurement noise on the uncertainty in the predictions. The dominant direction of uncertainty across all angles of attack arises primarily from the combined contributions of $\xi_1$ and $\xi_2$, indicating that measurement noise significantly affects the vorticity field predictions throughout the surrounding flow at a certain angle of attack. In contrast, the smaller uncertainty along the $\xi_3$ axis suggests that the pressure measurements are more sensitive to variations in the angle of attack. This behavior is influenced by the fact that sensor coordinates were stacked with the pressure measurements as inputs to the network in order to enhance the network's ability to distinguish between angles of attack. The alignment of the major axis of the ellipses, primarily tangential to the trajectory, suggests that the estimator's uncertainty about its predictions is most significant in the direction of preceding or subsequent snapshots. This behavior indicates that vorticity variations between consecutive snapshots lead to sensor measurement changes that remain within the uncertainty bounds, implying reduced measurements' sensitivity along this direction. 

\begin{figure*}[tb!]
\includegraphics[width=1.0\textwidth, trim=0 30 0 50, clip]{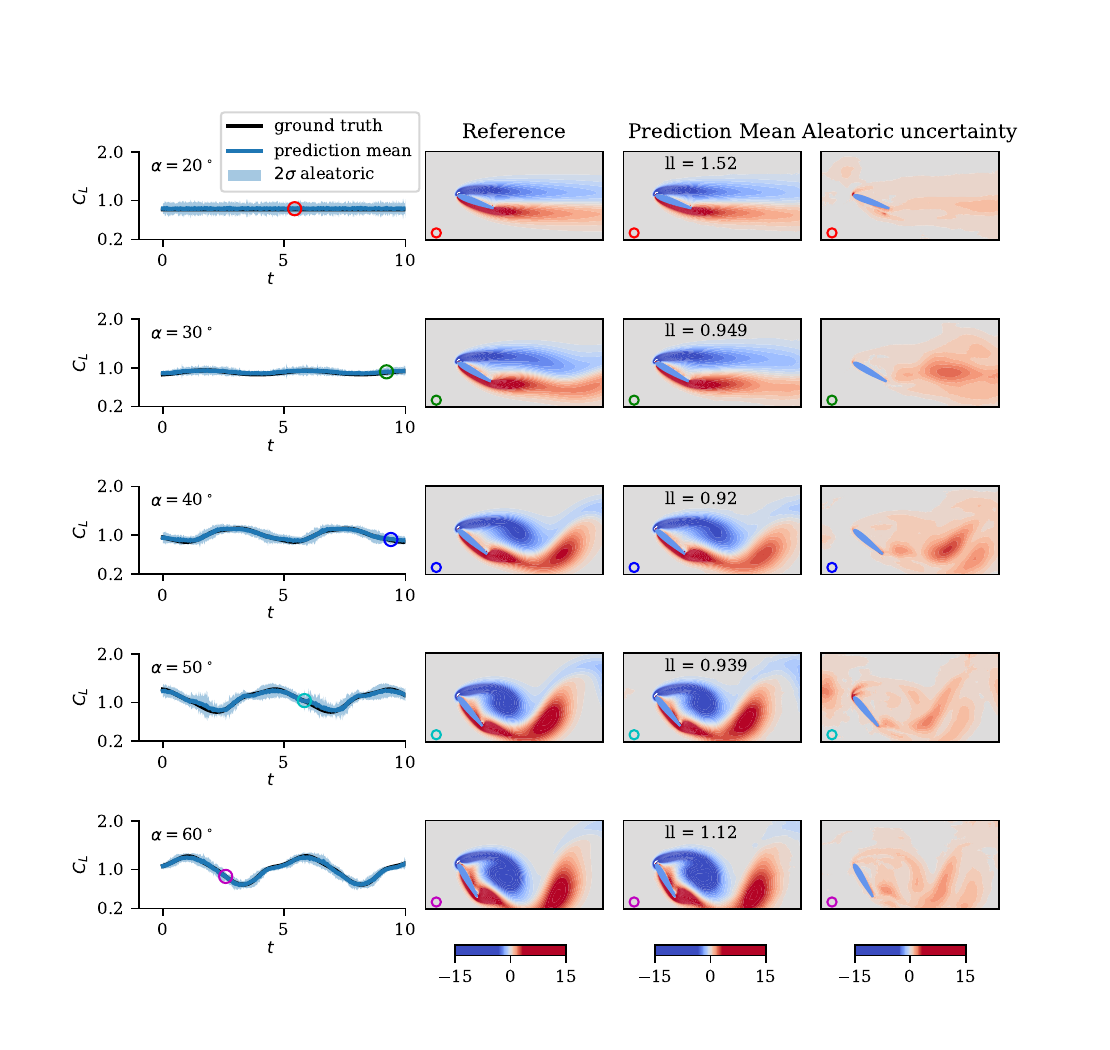}
\caption{\label{fig:aleatoric_uncertainty_vort_max_std_base} Aleatoric (data) uncertainty of five undisturbed cases due to measurement noise, represented by the predicted mean and two standard deviations for the lift and the vorticity fields. The left panels illustrate the evolution of the predicted lift coefficient alongside the ground truth for five different angles of attack. Symbols indicate the instants of maximum uncertainty in the predicted latent space, with the corresponding predicted vorticity field shown in the right panels. The far-right column presents the two standard deviations of the vorticity field. The term ``ll'' refers to the average pixel-wise log-likelihood of the predicted vorticity field computed following a Gaussian distribution.}
\end{figure*}

To compute the aleatoric uncertainty in the predicted flow field, we draw $M=100$ samples of the latent variables from the predicted probability distribution Eq.~\eqref{eq:pi_a} computed during the evaluation phase, and used these sampled latent variables to reconstruct samples of the vorticity and lift via Eq.~\eqref{eq:uncertainty-reconstruct}. The expected value and variance of the reconstructed samples indicate the characteristics of a Gaussian distribution. The left panels of Fig.~\ref{fig:aleatoric_uncertainty_vort_max_std_base} present the predicted lift history across five undisturbed cases, including the corresponding $95 \%$ confidence interval. The predictions show high accuracy across all angles of attack, except at $\alpha=50^\circ$. At this specific angle, even minor perturbations in the measurements can cause the predicted latent space samples to deviate, aligning with adjacent trajectories and compromising the robustness of the results. Despite this difficulty, the actual lift measurements consistently fall within the predicted uncertainty intervals.

The variance of the predicted samples fluctuates over time; in the heteroscedastic model, this variance is data-dependent, changing based on the input conditions. The instants when the variance in the predicted latent variables is maximum are indicated in Fig.~\ref{fig:aleatoric_uncertainty_x_lat_max_std} with thicker ellipses. These specific times are also marked in Fig.~\ref{fig:aleatoric_uncertainty_vort_max_std_base} using colored symbols. At these critical moments, when the predictions in the latent space exhibit considerable uncertainty, we further investigate the uncertainty in the predicted lift and vorticity fields as in Fig.~\ref{fig:aleatoric_uncertainty_vort_max_std_base}. In general, the instants when the latent variables' predictions show the maximum deviation from true trajectories do not align with the highest uncertainty in the lift and vorticity predictions. This is not surprising, given the non-linear nature of the decoder. The analysis shows that the greatest uncertainty in vorticity predictions at each snapshot occur in the regions of largest gradient of the vorticity field. This suggests two aspects of pressure-based estimation that deserve further study: first, the regions of large vorticity gradient also tend to coincide with dynamically important topological features, such as saddle points \citep{tu2022ftle}; also, these large-gradient regions are often associated with shear layers, where at higher Reynolds numbers, small-scale vortices and instabilities tend to form. While these smaller-scale structures introduce local turbulence, their impact on pressure readings is typically less significant than that of large-scale vortices in the wake and separation events near the airfoil itself.   

To assess the estimator's performance, we can evaluate the log-likelihood at the reference vorticity, given the predictive normal distribution of vorticity obtained from our samples. We average this log-likelihood across all pixels and show it as ``ll'' in the predicted mean panels. Interestingly, the predictions at all angles of attack perform reasonably well. The high positive log-likelihood values indicate a dense concentration of predicted values around the ground truth, suggesting that the model provides accurate overall predictions for all angles under study. 

\begin{figure*}[tb!]
\includegraphics[width=1.0\textwidth]{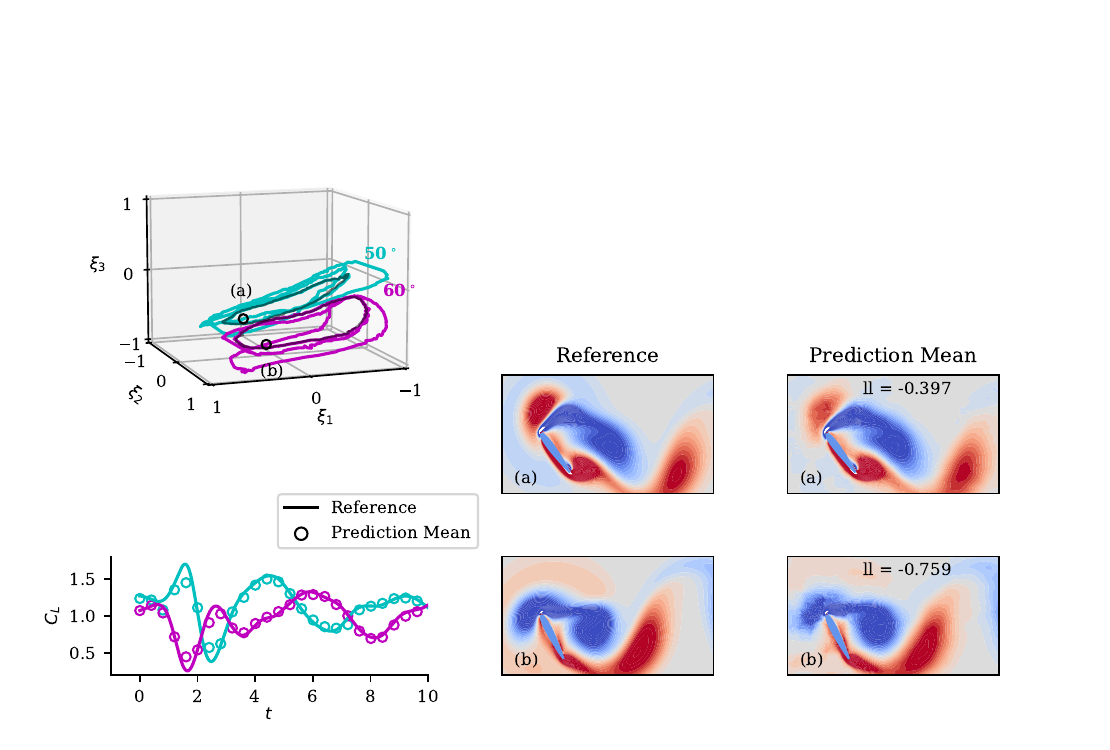}
\caption{\label{fig:vort_CL_xi_gust_aleatoric} Impact of a random gust on aerodynamics. The top panel displays the behavior of the true latent trajectory under two different conditions: ($\alpha=50^\circ$, $G=0.93$, $2R/c=0.68$, $y_o/c=0.09$); and ($\alpha=60^\circ$, $G=-0.98$, $2R/c=0.77$, $y_o/c=-0.26$). The corresponding undisturbed trajectories are illustrated by black closed curves. The predicted mean values from noisy measurements are compared with true values for both the lift history (lower left panel) and the vorticity field at two specific times (right panels), with the reported average pixel-wise log-likelihood, ``ll'', providing a quantitative measure of the prediction accuracy.}
\end{figure*}

\begin{figure*}[tb!]
\includegraphics[width=1.0\textwidth, trim=0 30 0 50, clip]{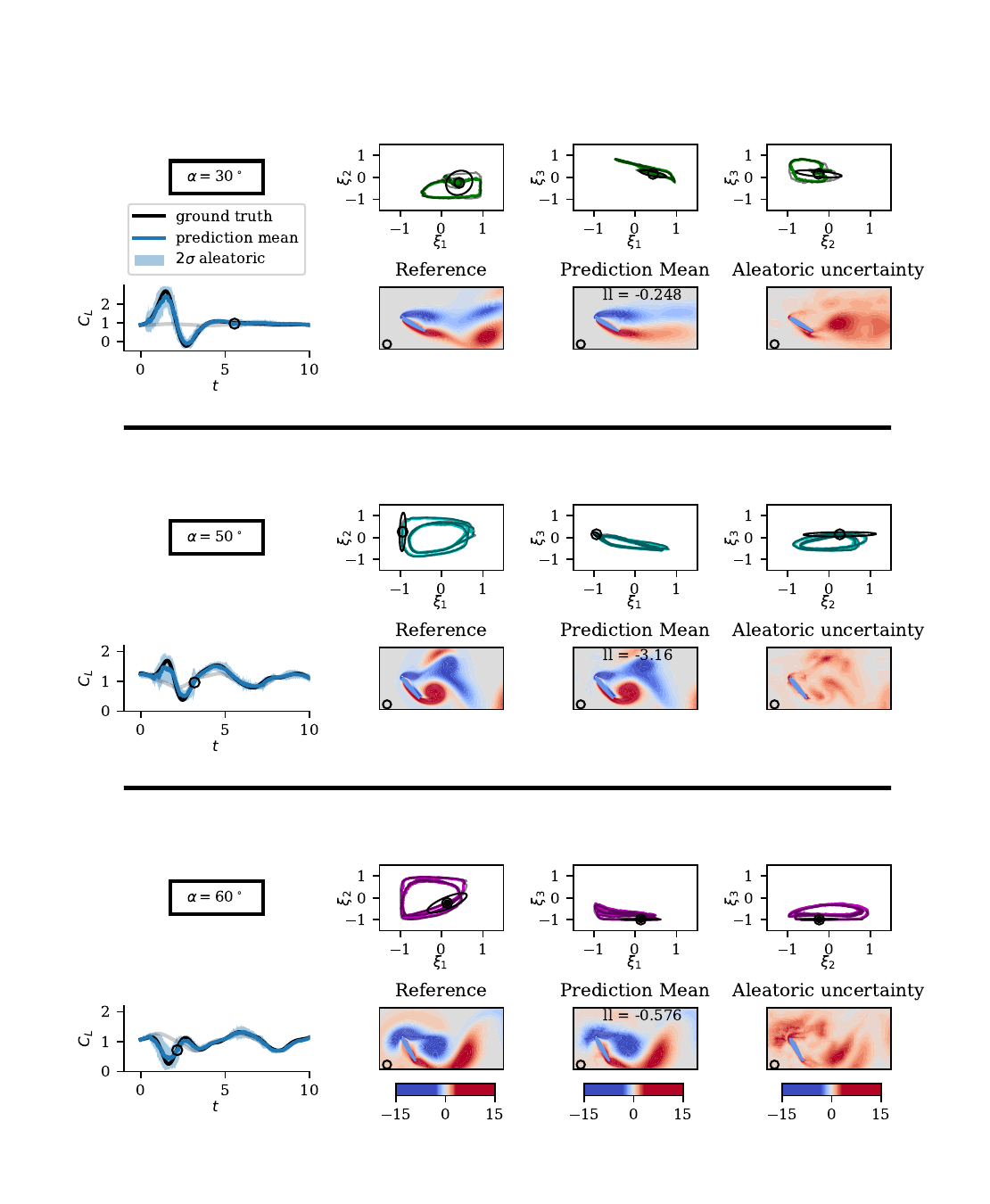}
\caption{\label{fig:aleatoric_uncertainty_vort_max_std_gust} Aleatoric uncertainty in gust-airfoil aerodynamics is illustrated with the predicted mean (solid-colored curves) and a $95 \%$ confidence interval for lift coefficient and vorticity field. From top to bottom, the conditions depicted are ($\alpha=30^\circ$, $G=0.93$, $2R/c=0.98$, $y_o/c=0.04$); ($\alpha=50^\circ$, $G=0.93$, $2R/c=0.68$, $y_o/c=0.09$); and ($\alpha=60^\circ$, $G=-0.98$, $2R/c=0.77$, $y_o/c=-0.26$). The light solid curves in the lift plots represent the corresponding undisturbed cases, providing a baseline for comparison. The grey solid orbits in the latent space illustrate the true trajectories. Symbols indicate the instants when deviations from the mean are at their peak in the latent space, with ellipses showing the eigenmodes of these deviations. The vorticity plots show predictions at these specific times, highlighting the impact of the gust on the flow field.}
\end{figure*}

We now shift our focus to gust-encounter scenarios. When a gust interacts with an airfoil, it influences both the effective angle of attack and the strength of vortices around the airfoil as the gust traverses the surface. This interaction significantly impacts the lift coefficient due to the altered aerodynamics. Figure~\ref{fig:vort_CL_xi_gust_aleatoric} illustrates how a disturbance affects the true latent variables as well as the true and predicted mean of the lift coefficient history and the vorticity field at a specific instant. In the latent space, the presence of a gust causes a marked deviation in the trajectory of latent variables along all three dimensions, highlighting the system's response to the perturbation. As the gust departs, the system gradually returns to its undisturbed periodic orbit, depicted by black closed curves, which is indicative of stable limit-cycle behavior as noted in previous studies \citep{raffoul2022advanced}. Moreover, as stated earlier, a positive disturbance (for the case of $\alpha=50^\circ$ in this figure) increases the lift while in the presence of a negative gust (for the case of $\alpha=60^\circ$), the lift initially drops. The plots of the lift coefficient and vorticity fields confirm that the predicted mean initially slightly deviates from the reference values when the gust hits the airfoil and gradually aligns closely with the reference values as the disturbance passes over it.

Figure~\ref{fig:aleatoric_uncertainty_vort_max_std_gust} presents aerodynamic predictions for an airfoil experiencing random disturbances at three distinct angles of attack: $\alpha \in \{30^\circ, 50^\circ, 60^\circ \}$. The comparison between the predicted means (with a $95 \%$ confidence interval) and reference data highlights several key aspects of the model's performance during gust encounters. The presence of gusts disrupts the normally periodic behavior of the latent space trajectories, pulling them away from their stable limit cycle. As the disturbance fades, however, the trajectories gradually return to their stable trajectory. This cyclical deviation and recovery showcase the model’s capability to capture the transient, complex nature of the gust-airfoil interaction. Despite the introduction of disturbances, the predicted means in the latent space maintain a strong alignment with the reference data at higher angles of attack. However, at lower angles, particularly in the $\xi_1 - \xi_2$ plane, the accuracy of the predictions diminishes, as evident in the larger covariance ellipse in that plane. Nevertheless, even with this degradation, the true trajectory remains within the uncertainty bounds, suggesting that the model still captures the underlying flow dynamics despite increased uncertainty. The predicted lift evolution closely tracks the reference values, with a minor deviation observed at the peak due to the passage of the gust. A key insight from the figure is the model’s capacity to effectively capture the heightened uncertainty during gust-airfoil interactions. As gusts induce more variability in the flow, the model appropriately widens the uncertainty bounds, reflecting a reduced confidence in the predictions. This expanded confidence interval is vital, as it ensures that, despite the added complexity of gust disturbances, the true values stay within the predicted uncertainty range.

A notable observation is the difference between disturbed and undisturbed cases: the uncertainty of the predicted latent variables during gust encounters is consistently larger than in the undisturbed flow at similar time instants. This reflects how the model captures the added complexity and instability introduced by the gusts. Additionally, comparing the uncertainty ellipses in the latent space for disturbed cases with undisturbed cases (Fig.~\ref{fig:aleatoric_uncertainty_x_lat_max_std} and Fig.~\ref{fig:aleatoric_uncertainty_vort_max_std_gust}) shows that the greatest uncertainty in the dominant directions during gusts is noticeably larger. The moment of peak uncertainty in the latent space aligns with either the gust's approach to the airfoil or its direct interaction with it. As in undisturbed cases, the pressure measurements during disturbed aerodynamics are observed to display low sensitivity to changes in the flow structure across consecutive snapshots.
At the point of maximum uncertainty in the latent space, the corresponding predicted vorticity field is shown for further analysis. The elevated uncertainty regions are predominantly located within the wake, particularly around the shear layers and high vorticity gradient regions, as well as in the vicinity of the gust. These areas are characterized by more complex flow dynamics, driven by unsteady aerodynamic effects that introduce greater variability and challenge the model's ability to provide accurate predictions. 
The reduced performance of the trained estimator in predicting the mean vorticity field at lower angles of attack, specifically at $\alpha=30^\circ$, is reflected in the deviations of the predicted latent vector from the true trajectory in the $\xi_1-\xi_2$ plane. These discrepancies occur in regions of the flow far from the airfoil, where the variations in vorticity have a negligible impact on the surface pressure measurements within the associated uncertainty bounds.

\begin{figure*}[tb!]
\includegraphics[width=1.0\textwidth]{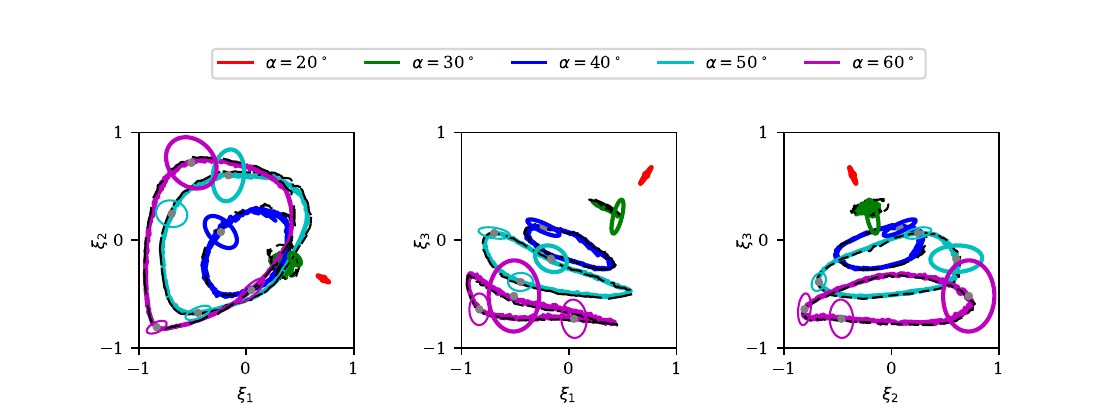}
\caption{\label{fig:epistemic_uncertainty_x_lat_max_std} Epistemic (model) uncertainty of estimation of five undisturbed cases. Predicted mean with $95 \%$ confidence ellipses of latent variables at a small number of instants for five undisturbed cases are shown. The solid-colored curves represent the mean of $\hat{\bo{\mu}}$, while the dashed black curves indicate the true trajectories extracted from the lift-augmented autoencoder. Thicker ellipses correspond to periods of maximum uncertainty.}
\end{figure*}

\subsubsection{Epistemic Uncertainty}
Epistemic uncertainty, stemming from incomplete knowledge about the trained DL model, is crucial in data-driven studies of gust-encounter aerodynamics. Unlike aleatoric uncertainty, epistemic uncertainty can be reduced through better models and additional data. To quantify this, in Section~\ref{sec:quantify-uncertainty} we described a probabilistic approach using MC dropout to sample from the model's weights during inference. We use the samples generated in the last section to estimate the model's epistemic probability density function with Eq.~\eqref{eq:pi_e}, capturing both mean and covariance information through the statistics of the output samples $\{\hat{\bo{\mu}}_k\}_{k=1}^T$. This method allows us to estimate the reliability of our predictions and refine our understanding of the transient flow fields and aerodynamic loads. In this section, we will detail our results, highlighting how MC dropout helps assess unreliability in the trained model.

Figure~\ref{fig:epistemic_uncertainty_x_lat_max_std} illustrates the estimated means of the latent variables alongside their $95 \%$ confidence ellipses across five undisturbed cases. This figure highlights the model's ability to compress and predict the underlying flow dynamics accurately, despite variations in the fitted model. The ellipses, representing epistemic uncertainty, highlight the degree of imprecision arising from the model's inherent knowledge gaps or limitations due to insufficient data. The major axes of these ellipses are predominantly perpendicular to the trajectories, reflecting the scarcity of training data away from the primary trajectory paths. These broader uncertainty regions reveal the model's performance in terms of the epistemic uncertainty, especially when it encounters conditions that differ from what it has been trained on, reflecting the inherent limitations of the network or data.
\begin{figure*}[tb!]
\includegraphics[width=1.0\textwidth, trim=0 30 0 30, clip]{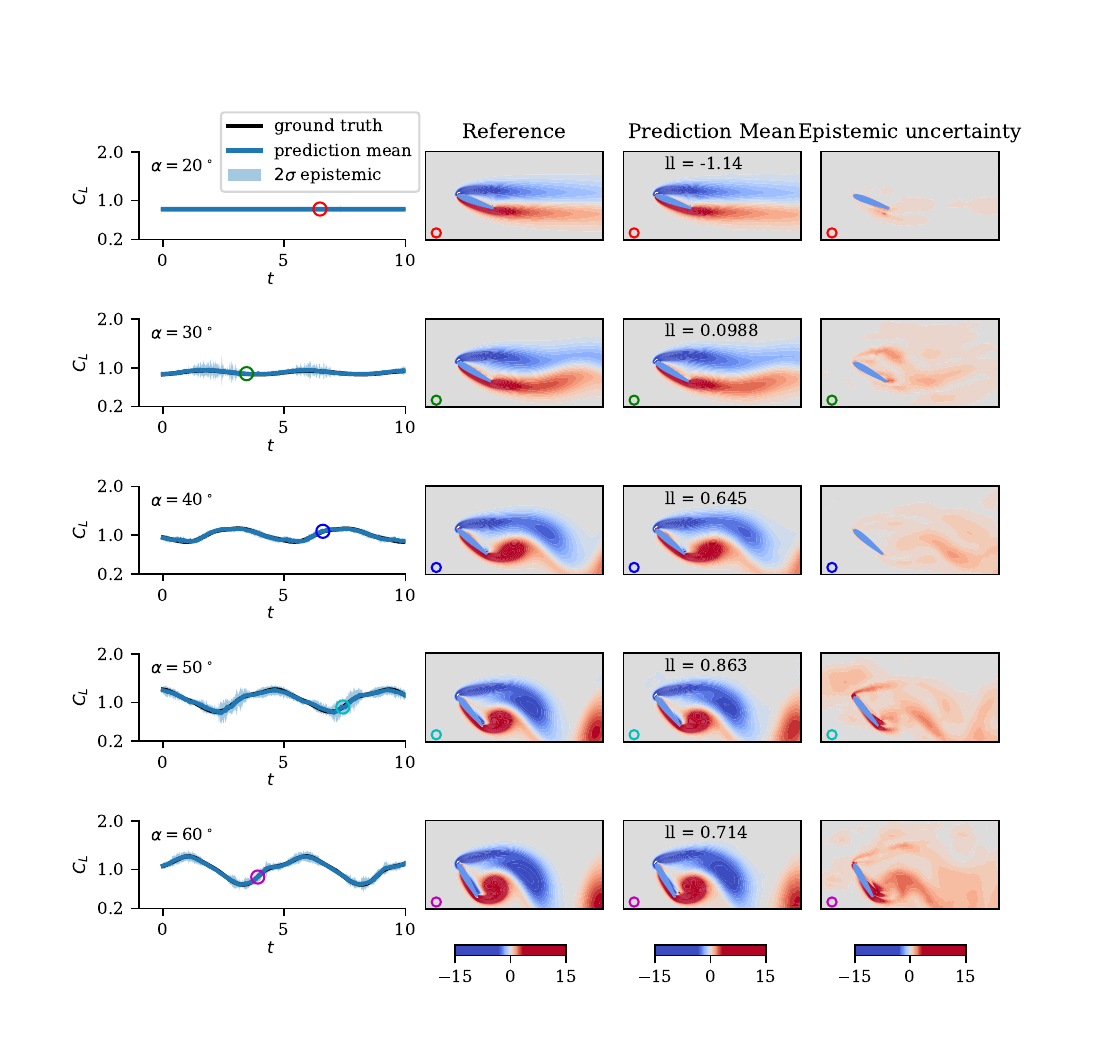}
\caption{\label{fig:epistemic_uncertainty_vort_max_std_base} Epistemic (model) uncertainty of five undisturbed cases, represented by the predicted mean and two standard deviations for the lift and vorticity fields. The left panels illustrate the evolution of the predicted lift coefficient alongside the ground truth for five different angles of attack. Symbols indicate the instants of maximum uncertainty in the predicted latent space, with the corresponding predicted vorticity field shown in the right panels. The far-right column presents the two standard deviations of the vorticity field.}
\end{figure*}

To assess the epistemic uncertainty in lift and vorticity, we draw $M=100$ samples of the latent variables from the predicted probability distribution Eq.~\eqref{eq:pi_e} and reconstruct samples of vorticity via the procedure described by Eq.~\eqref{eq:uncertainty-reconstruct}. The expected value and variance of the reconstructed samples indicate the characteristics of a Gaussian distribution.
Figure~\ref{fig:epistemic_uncertainty_vort_max_std_base} presents the mean predictions along with their $95 \%$ confidence intervals, within which the true lift is contained. The symbols along the lift curves in Fig.~\ref{fig:epistemic_uncertainty_vort_max_std_base} represent the snapshots when the predicted latent space samples exhibit the greatest spread around the mean (indicated by thicker ellipses in Fig.~\ref{fig:epistemic_uncertainty_x_lat_max_std}). The corresponding vorticity fields at these instants are displayed on the right, highlighting regions of maximum uncertainty in the wake. The model demonstrates significant epistemic uncertainty in regions of complex flow dynamics, particularly within regions of large vorticity gradient, such as shear layers and between vortices, which further suggests that the model would struggle most in regions of small-scale flow interactions at higher Reynolds numbers. 

At lower angles such as $\alpha=20^\circ$, the flow remains relatively steady and attached to the airfoil surface, making it easier for the model to predict the aerodynamic characteristics. At higher angles, the flow becomes fully unsteady, but the model has learned to handle these consistently unsteady patterns. The model exhibits lower confidence in its predictions in regions where the training data is sparse or underrepresented. This reduced certainty likely stems from the model's limited exposure to the dynamics in these areas during training, making it more challenging to accurately capture and generalize the complex flow behavior. As a result, the model's predictive accuracy decreases, and uncertainty increases in these regions. The log-likelihood per pixel is reported for the estimated vorticity field in each case, providing further insight into the model's performance. The negative low log-likelihood for $\alpha=20^\circ$ is a result of the steep penalty this metric applies to deviations from the mean when the variance is small. 

The analysis in Fig.~\ref{fig:epistemic_uncertainty_x_lat_max_std} showed that the dominant uncertainty spans all three directions within the latent space across all cases, in contrast to the case of aleatoric uncertainty, in which uncertainty was mostly limited to the $\xi_1$ and $\xi_2$ directions. The $\xi_3$ uncertainty reflects uncertainty in the angle of attack. This is particularly notable in the covariance ellipses for $\alpha \in \{40^\circ, 50^\circ, 60^\circ \}$, which overlap with the mean trajectories of neighboring angles in the latent space. This multidirectional uncertainty suggests that randomness in the model weights gives rise to large variations in the vorticity predictions not only far from the airfoil (where uncertainty in $\xi_1$ and $\xi_2$ plays a prominent role, as in the case of aleatoric uncertainty), but also adjacent to it (where angle of attack uncertainty is most influential). Indeed, the uncertainty plots of the predicted vorticity field presented in Fig.~\ref{fig:epistemic_uncertainty_vort_max_std_base} illustrate the estimator's confusion regarding the vorticity field adjacent to the airfoil at the specified angles of attack, moreso than in the case of aleatoric uncertainty. In particular, at the lower angles of attack, $\alpha \in \{20^\circ, 30^\circ \}$, in which uncertainty in $\xi_3$ is lower, the model exhibits less uncertainty about the vorticity field close to the airfoil.

\begin{figure*}[tb!]
\includegraphics[width=1.0\textwidth, trim=0 30 0 50, clip]{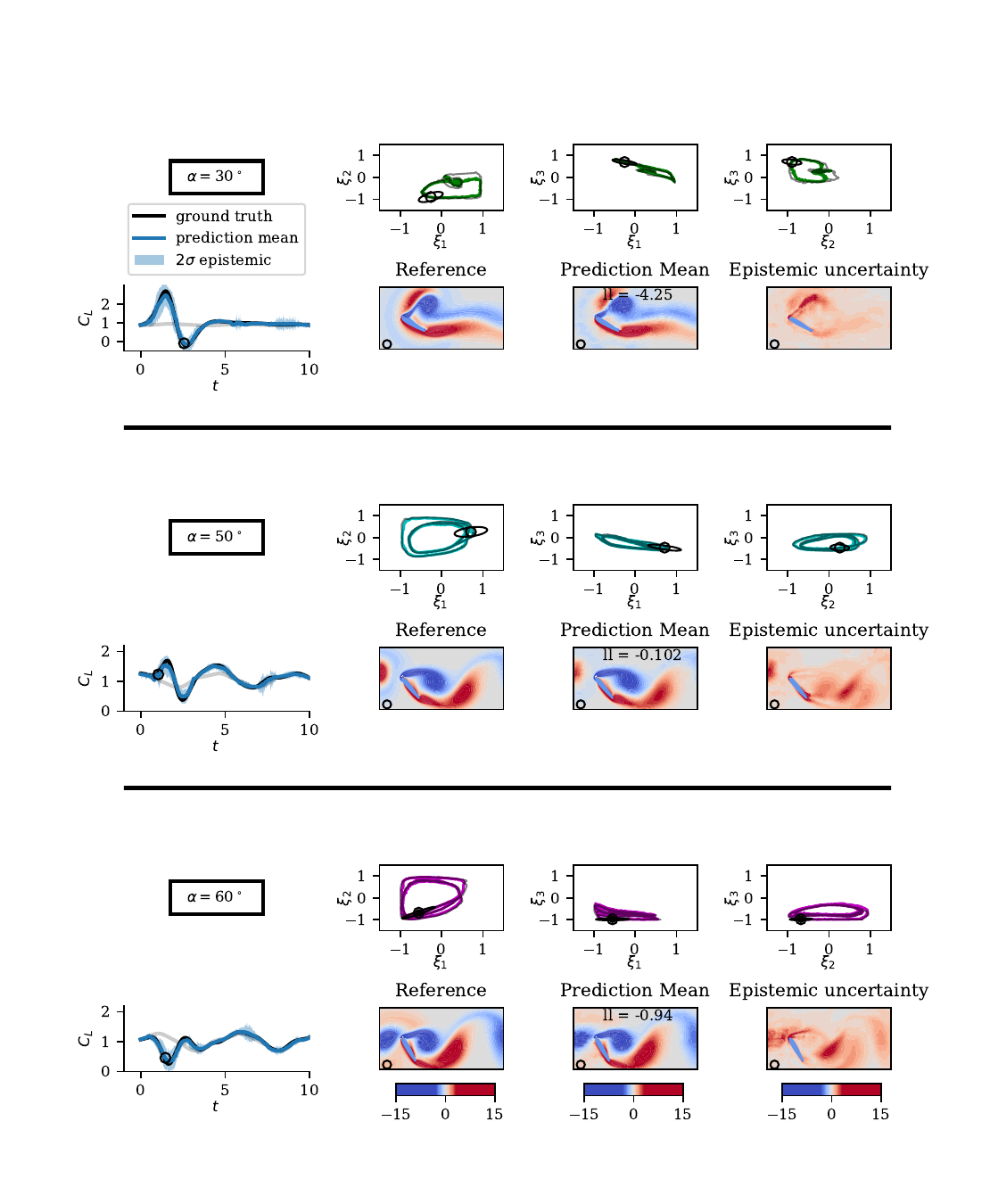}
\caption{\label{fig:epistemic_uncertainty_vort_max_std_gust} Epistemic uncertainty in gust-airfoil aerodynamics is illustrated with the predicted mean (solid-colored curves) and a $95 \%$ confidence interval for lift coefficient and vorticity field. From top to bottom, the conditions depicted are ($\alpha=30^\circ$, $G=0.93$, $2R/c=0.98$, $y_o/c=0.04$); ($\alpha=50^\circ$, $G=0.93$, $2R/c=0.68$, $y_o/c=0.09$); and ($\alpha=60^\circ$, $G=-0.98$, $2R/c=0.77$, $y_o/c=-0.26$). The light solid curves in the lift plots represent the corresponding undisturbed cases, providing a baseline for comparison. The grey solid orbits in the latent space illustrate the true trajectories. Symbols indicate the instants of greatest deviation from the mean in the latent space, with ellipses showing the eigenmodes of these deviations. The vorticity plots show predictions at these specific times, highlighting the impact of the gust on the flow field.}
\end{figure*}

In Fig.~\ref{fig:epistemic_uncertainty_vort_max_std_gust} we investigate the epistemic uncertainty of the model's performance in gusty aerodynamics under the same conditions as in Fig.~\ref{fig:aleatoric_uncertainty_vort_max_std_gust}. The predicted mean lift closely aligns with the ground truth in all cases. Unlike aleatoric uncertainty, the model demonstrates greater uncertainty in its latent space predictions for base cases compared to disturbed cases. This indicates that the trained model is less capable of distinguishing undisturbed aerodynamics compared to disturbed flow conditions. The instant when the norm of the uncertainty ellipse is maximum is depicted in the latent space. This instant is primarily linked to the gust's approach or interaction with the airfoil. Notably, the point of maximum uncertainty in the latent variables does not always coincide with the peak variance in lift or vorticity due to the non-linearity of the decoder. Similar to aleatoric uncertainty modeling, the model shows the highest uncertainty during gust-airfoil interactions, especially around the first lift peak, due to the highly complex aerodynamic behavior. The mean of the estimated vorticity clearly indicates the disturbance, and the model also identifies it through a region of heightened uncertainty in its predictions. When a gust approaches and interacts with the wake, the model shows increased uncertainty not only at vortex boundaries but also at the disturbance location, highlighting its awareness of the complex flow interactions and the presence of a gust.

\section{Conclusion}\label{conclusion}
This study has developed a deep-learning approach for reconstructing gust-encounter flow fields and lift coefficients and their respective uncertainties using sparse and noisy surface pressure measurements. We have demonstrated the approach on undisturbed and disturbed two-dimensional low Reynolds number flow about an airfoil at a variety of angles of attack. By employing a non-linear lift-augmented autoencoder, we effectively reduced the high-dimensional flow data to three latent components, greatly minimizing the computational demands for sensor-based estimation. Our investigation has offered an in-depth analysis of how sensors dynamically respond to gust-airfoil interactions, revealing the transient importance of sensors in different positions on the airfoil surface during gust encounters.

It was observed that during the interaction of a gust, regardless of its polarity, with the negative leading-edge vortex, sensors located on the suction surface of the airfoil consistently captured critical information about the localized flow changes. This dominance persisted as the gust traversed up to a distance of approximately one chord length from the airfoil. Beyond this point, the influence of the gust shifted, with sensors on the pressure side becoming the primary indicators of flow variations. Once the gust moved far away from the airfoil, the transient sensor responses gradually returned to their periodic behavior, characteristic of the undisturbed flow conditions.

To effectively manage the uncertainties inherent in deep learning models, we employed strategies to model both aleatoric and epistemic uncertainties. Aleatoric uncertainty, arising from sensor measurement noise, was addressed using a heteroscedastic loss to predict the parameters of a multivariate normal distribution in the latent space. To enhance robustness, we introduced noise into the measurements during training, ensuring the model's ability to produce reliable predictions despite variability in sensor readings. The directions of the greatest uncertainty in the covariance ellipsoid in the latent space were found to be nearly tangent to the latent variable trajectories, indicating that the estimator’s uncertainty is most pronounced in the direction of preceding or subsequent snapshots. Additionally, the pressure measurements were found to be highly sensitive to the airfoil’s angle of attack.

Epistemic uncertainty, stemming from the model's dearth of training, was tackled using Monte Carlo dropout to capture uncertainty in the model itself. Unlike aleatoric uncertainty, epistemic uncertainty is most significant in directions perpendicular to the latent variable trajectories, which corresponds to regions with less training data. The stochasticity of both the model and the pressure readings results in high uncertainty when predicting compressed representation of the flow, lift, and vorticity fields, particularly during gust-airfoil interactions. For both types of uncertainty, the highest levels of uncertainty in the predicted vorticity were observed at the vortex boundaries, where small-scale vortices and shear instabilities tend to develop at higher Reynolds numbers. We hypothesize that these small-scale structures would likely have a minimal impact on pressure sensor readings and the trained model compared to the larger-scale flow phenomena that emerge from flow separation and subsequent vortex shedding around the airfoil.

The results have demonstrated the effectiveness of our approach in accurately quantifying uncertainties within complex aerodynamic environments, all while maintaining computational efficiency. This highlights its potential to significantly enhance sensor-based flow field estimations. The uncertainty analysis presented highlights the vital importance of addressing both aleatoric and epistemic uncertainties to achieve robust and reliable predictions, particularly in complex aerodynamic environments. This uncertainty quantification can be leveraged to improve model performance. For example, by strategically employing active learning, data sampling can be concentrated in the dominant directions of uncertainty, thereby efficiently reducing model uncertainty and enhancing predictive accuracy. This approach not only strengthens the reliability of the model but also optimizes resource utilization in data acquisition.

There are a number of other aspects of this study to pursue further. We have observed the largest uncertainty in regions of high vorticity gradients. These regions often coincide with kinematically important features such as saddle points, and it would be useful to further explore this connection more deeply \citep{tu2022ftle}. Also, while the current study employed synthetic sensor data with added Gaussian noise, the model's performance should be validated with real-time sensor measurements and experimental flow data. Moreover, the versatility of this approach allows for its application beyond aerodynamics, potentially broadening its impact across various domains.

\appendix
\renewcommand{\thetable}{A\arabic{table}}
\section{Appendix}
The architecture of the neural network used to map surface measurements to the low-dimensional latent space $\mathcal{F}_p$ is detailed in Table.~\ref{tab:network_architecture}. In this network, to adopt MC-dropout approach, dropout layers are strategically incorporated after each dense layer. During the training and inference phases, these dropout layers are activated to capture aleatoric and epistemic uncertainties by introducing randomness in the network weights.
\begin{table}[h!]
  \caption{Structure of the sensor-based network employed in the present study which maps the pressure measurements to the latent variables.}
  \centering
  \begin{tabular}{|c|c|c|}
    \hline
    \textbf{Layer} & \textbf{Data size} & \textbf{Dropout rate} \\
    \hline
    \hline
    Input   & (33)  & - \\
    \hline
    Fully connected   & (64)  & - \\
    \hline
    Dropout   & -  & $(1-p)$ \\
    \hline
    Fully connected   & (128)  & - \\
    \hline
    Dropout   & -  & $(1-p)$ \\
    \hline
    Fully connected   & (256)  & - \\
    \hline
    Dropout   & -  & $(1-p)$ \\
    \hline
    Fully connected   & (512)  & - \\
    \hline
    Dropout   & -  & $(1-p)$ \\
    \hline
    Fully connected   & (256)  & - \\
    \hline
    Dropout   & -  & $(1-p)$ \\
    \hline
    Fully connected   & (128)  & - \\
    \hline
    Dropout   & -  & $(1-p)$ \\
    \hline
    Fully connected   & (64)  & - \\
    \hline
    Dropout   & -  & $(1-p)$ \\
    \hline
    Fully connected (latent vector mean)   & (3) & - \\
    \hline
    Fully connected (elements of a lower-triangular matrix)   & (6) & - \\
    \hline
  \end{tabular}
  \label{tab:network_architecture}
\end{table}

\bibliographystyle{apalike}
\bibliography{refs}

\end{document}